\documentclass[twoside,11pt]{article}

\usepackage{jmlr2e}

\usepackage{hyperref}
\usepackage{url}
\usepackage{graphicx} 
\usepackage{algorithm} 
\usepackage{algorithmic} 
\usepackage{amsmath,amssymb}
\usepackage{wrapfig}
\usepackage{comment}

\newtheorem{Definition}{{\bf Definition}}[section]
\newtheorem{Proposition}[Definition]{{\bf Proposition}}

\newtheorem{Theorem}[Definition]{{\bf Theorem}}
\newtheorem{Lemma}[Definition]{{\bf Lemma}}

\newtheorem{Example}[Definition]{{\bf Example}}

\jmlrheading{17}{2016}{1-28}{3/14; Revised 5/16}{9/16}{Yu Nishiyama and Kenji Fukumizu}

\ShortHeadings{Characteristic Kernels and Infinitely Divisible Distributions}{Nishiyama and Fukumizu}
\firstpageno{1}

\begin{document}

\title{Characteristic Kernels and Infinitely Divisible Distributions}

\author{\name Yu Nishiyama \email ynishiyam@gmail.com \\
       \addr The University of Electro-Communications \\
       1-5-1 Chofugaoka, Chofu, Tokyo 182-8585, Japan
       \AND
       \name Kenji Fukumizu \email fukumizu@ism.ac.jp \\
       \addr The Institute of Statistical Mathematics \\
       10-3 Midori-cho, Tachikawa, Tokyo 190-8562, Japan}

\editor{Ingo Steinwart}

\maketitle

\begin{abstract}
We connect shift-invariant characteristic kernels to infinitely divisible distributions on $\mathbb{R}^{d}$.
Characteristic kernels play an important role in machine learning applications with their kernel means to distinguish any two probability measures.
The contribution of this paper is twofold.
First, we show, using the L\'evy--Khintchine formula, that any shift-invariant kernel given by a bounded, continuous, and symmetric probability density function (pdf) of an infinitely divisible distribution on $\mathbb{R}^d$ is characteristic. 
We mention some closure properties of such characteristic kernels under addition, pointwise product, and convolution.
Second, in developing various kernel mean algorithms, it is fundamental to compute the following values: (i) kernel mean values $m_P(x)$, $x \in \mathcal{X}$, and (ii) kernel mean RKHS inner products ${\left\langle m_P, m_Q \right\rangle _{\mathcal{H}}}$, for probability measures $P, Q$.
If $P, Q$, and kernel $k$ are Gaussians, then the computation of (i) and (ii) results in Gaussian pdfs that are tractable. We generalize this Gaussian combination to more general cases in the class of infinitely divisible distributions. 
We then introduce a {\it conjugate} kernel and a {\it convolution trick}, so that the above (i) and (ii) have the same pdf form, expecting tractable computation at least in some cases. 
As specific instances, we explore $\alpha$-stable distributions and a rich class of generalized hyperbolic distributions, where the Laplace, Cauchy, and Student's $t$ distributions are included.
\end{abstract}


\section{Introduction} \label{sec:Introduction}
Let $(\mathcal{X}, \mathcal{B}(\mathcal{X}))$ be a measurable space and $\mathcal{M}_1(\mathcal{X})$ be the set of probability measures.
Let $\mathcal{H}$ be the real-valued reproducing kernel Hilbert space (RKHS) associated with a bounded and measurable positive-definite (p.d.) kernel $k: \mathcal{X} \times \mathcal{X} \rightarrow \mathbb{R}$.
In machine learning, kernel methods provide a technique for developing nonlinear algorithms, by mapping data $X_1, \cdots, X_n$ in $\mathcal{X}$ to higher- or infinite-dimensional RKHS functions $k(\cdot, X_1), \ldots, k(\cdot, X_n)$ in $\mathcal{H}$ \citep{LearningwithKernels2002, SupportVectorMachines2008}.

Recently, an RKHS representation of a probability measure $P \in \mathcal{M}_1(\mathcal{X})$, called kernel mean, $m_P := \mathbb{E}_{X \sim P}[k(\cdot, X)] \in \mathcal{H}$ \citep{Smola_etal_ALT2007, KernelBayes'Rule_BayesianInferencewithPositiveDefiniteKernels}, or equivalently,
\begin{equation}\label{eq:kernel_mean_integral}
m_P(x)=\int k(x,y)dP(y), \hspace{2mm} x \in \mathcal{X}
\end{equation}
has been used to handle probability measures in RKHSs.
The kernel mean enables us to introduce a similarity and distance between two probability measures $P, Q \in \mathcal{M}_1(\mathcal{X})$, via the RKHS inner product ${\left\langle m_P, m_Q \right\rangle _{\mathcal{H}}}$ and the norm $|| m_P -m_Q ||_{\mathcal{H}}$, respectively.  Using these quantities, different authors have proposed many algorithms, including density estimations \citep{Smola_etal_ALT2007, TailoringICML2008, McCalman2013}, hypothesis tests (\citealt{DBLP:journals/jmlr/GrettonBRSS12}, \citealt{Gretton2008independencetest}, \citealt{NIPS2007_559}), kernel Bayesian inference  (\citealt{Song+al:icml2010hilbert}, \citealt{Song:2010fk}, \citealt{DBLP:journals/jmlr/SongGBLG11}, \citealt{KernelBayes'Rule_BayesianInferencewithPositiveDefiniteKernels},  \citealt{KernelEmbeddingofConditionalDistributions}, \citealt{KernelMonteCarlo2016}, \citealt{KernelBayesSmoothingAISTATS2016}), classification \citep{NIPS2012_0015}, dimension reduction \citep{NIPS2012_1036}, and reinforcement learning (\citealt{DBLP:journals/corr/abs-1206-4655}, \citealt{DBLP:conf/uai/NishiyamaBGF12}, \citealt{PathIntegralControlbyReproducingKernelHilbertSpaceEmbedding2013},  \citealt{HilbertSpaceEmbeddingsofPredictiveStateRepresentations}).

In these applications, the characteristic property of a p.d.~kernel $k$ is important: a p.d.~kernel is said to be {\em characteristic} if any two probability measures $P, Q \in \mathcal{M}_1(\mathcal{X})$ can be distinguished by their kernel means $m_{P}, m_{Q} \in \mathcal{H}$ \citep{DBLP:journals/jmlr/FukumizuBJ03, Sriperumbudur_JMLR2010, Sriperumbudur_JMLR2011}.
For a continuous, bounded, and shift-invariant p.d.~kernel on $\mathbb{R}^d$ with $k(x,y) = \kappa(x-y)$, a necessary and sufficient condition for the kernel to be characteristic is known via the Bochner theorem \citep[Theorem 9]{Sriperumbudur_JMLR2010}.

As the first contribution of this paper, we show, using the L\'evy--Khintchine formula \citep{Sato1999, InfinitelyDivisibleRealLine2004, Applebaum2009}, that if $\kappa$ is a continuous, bounded, and symmetric pdf of an infinitely divisible distribution $P$ on $\mathbb{R}^d$, then $k$ is a characteristic p.d.~kernel.  We call such kernels {\it convolutionally infinitely divisible} (CID) kernels. Examples of CID kernels are given in Example \ref{ex:symmetricinfinitedivisibledistribution}.
In addition, we note some closure properties of the CID kernels with respect to addition, pointwise product, and convolution.

To describe the second contribution, we briefly explain what is essentially computed in kernel mean algorithms. 
In general kernel methods, the following computations are fundamental:
\begin{itemize}
\item[(i)] RKHS function values: $f(x)$ for $f \in \mathcal{H}$, $x \in \mathcal{X}$,
\item[(ii)] RKHS inner products: ${\left\langle f, g \right\rangle _{\mathcal{H}}}$, $f,g \in \mathcal{H}$.
\end{itemize}
If $f \in \mathcal{H}$ is represented by $f := \sum_{i=1}^{n} w_i k(\cdot, X_i)$, $w \in \mathbb{R}^n$, then the function value (i) $f(x) = \sum_{i=1}^{n} w_i k(x, X_i)$ reduces to the evaluation of the kernel $k(x,y)$. Similarly, if two RKHS functions $f,g \in \mathcal{H}$ are both represented by $f := \sum_{i=1}^{n} w_i k(\cdot, X_i)$ and $g := \sum_{j=1}^{l} \tilde w_j k(\cdot, \tilde X_j)$, respectively, then the inner product (ii) ${\left\langle f, g \right\rangle _{\mathcal{H}}} = \sum_{i=1}^{n} \sum_{j=1}^{l} w_i \tilde w_j k(X_i, \tilde X_j)$ reduces to the evaluation of the kernel $k(x,y)$, which is so-called the {\it kernel trick} ${\langle k(\cdot, x), k(\cdot, y) \rangle _{\mathcal{H}}} = k(x, y)$.   

We consider a more general case in which $f, g \in \mathcal{H}$ are represented by $f := \sum_{i=1}^{n} w_i m_{P_i}$ and $g := \sum_{j=1}^{l} \tilde w_j m_{Q_j}$, respectively, where $\{m_{P_i} \}, \{ m_{Q_j} \} \subset \mathcal{H}$ are kernel means of probability measures $\{P_i\}, \{Q_j\} \subset \mathcal{M}_1(\mathcal{X})$. Kernel algorithms involving kernel means use this type of RKHS functions explicitly or implicitly.  
If $\{P_i\}, \{Q_j\}$ are delta measures $\{ \delta_{X_i} \}, \{ \delta_{\tilde X_j} \}$,\footnote{A probability measure $\delta_x(\cdot)$, $x \in \mathcal{X}$ is a delta measure; if $x \in B$, then $\delta_x(B)=1$; otherwise, $\delta_x(B)=0$ for $B \in \mathcal{B}(\mathcal{X})$.} then these functions are specialized to the above kernel trick case, where $m_{\delta_x} =k(\cdot, x)$. The computation of (i) $f(x) = \sum_{i=1}^{n} w_i m_{P_i}(x)$ and (ii) ${\left\langle f, g \right\rangle _{\mathcal{H}}} = \sum_{i=1}^{n} \sum_{j=1}^{l} w_i \tilde w_j {\langle m_{P_i}, m_{Q_j} \rangle _{\mathcal{H}}}$ requires the following kernel mean evaluations:\footnote{If kernel means $m_{P}, m_{Q}$ are also both expressed by a weighted sum, $m_{P} := \sum_{i=1}^{n_P} \eta_i k(\cdot, \dot X_i)$ and $m_{Q} := \sum_{i=1}^{n_Q} \tilde \eta_i k(\cdot, \ddot X_i)$, $\{ \dot X_i \}, \{ \ddot X_i\} \subset \mathcal{X}$, then the computation also reduces to the above kernel trick case.}

\begin{itemize}
\item[(iii)] kernel mean values: $m_{P}(x)$ for $P \in \mathcal{M}_{1}(\mathcal{X})$, $x \in \mathcal{X}$,
\item[(iv)] kernel mean inner products: ${\left\langle m_P, m_Q \right\rangle _{\mathcal{H}}}$, $P,Q \in \mathcal{M}_{1} (\mathcal{X})$.
\end{itemize}
Note that the kernel mean value (\ref{eq:kernel_mean_integral}) and the kernel mean inner product ${\left\langle m_P, m_Q \right\rangle _{\mathcal{H}}}=\int k(x,y)dP(x)dQ(y)$ involve an integral, and their rigorous computation is not tractable in general. 

The second contribution of this paper is to provide some classes of p.d. kernels and parametric models $P, Q \in \mathcal{P}_{\Theta} := \{{P}_{\theta}| \theta \in \Theta \}$ such that the kernel computation of (iii) and (iv) can be reduced to a kernel evaluation, where tractable computation is considered. For a shift-invariant kernel $k(x,y) = \kappa(x-y)$, $x,y \in \mathbb{R}^d$ on $\mathbb{R}^d$, as shown in Lemma \ref{lm:kernelmeanprobabilitydensityfunction}, the computation of (iii) and (iv) reduces to the following convolution:
\begin{itemize}
\item[(iii)'] kernel mean values: $m_{P}(x) = (\kappa * P)(x)$, 
\item[(iv)'] kernel mean inner products: ${\left\langle m_P, m_Q \right\rangle _{\mathcal{H}}} = (\kappa * \tilde P * Q)(0) = (\kappa * P * \tilde Q)(0)$, 
\end{itemize}
where $\tilde P$ and $\tilde Q$ are the dual of $P$ and $Q$, respectively.\footnote{A probability measure $\tilde{P} \in \mathcal{M}_1(\mathbb{R}^d)$ is called a \textit{dual} of $P \in \mathcal{M}_1(\mathbb{R}^d)$ if $\tilde{P}(B)=P(-B)$ for every $B \in \mathcal{B}(\mathbb{R}^{d})$, where $-B:=\{-x:x \in B \}$ \cite[p.8]{Sato1999}}
This convolution representation motivates us to explore a set of parametric distributions $\mathcal{P}_{\Theta}$ that is closed under convolution, namely, a convolution semigroup $(\mathcal{P}_{\Theta}, *) \subset \mathcal{M}_1(\mathbb{R}^{d})$, where $\kappa$ is a density function in $\mathcal{P}_{\Theta}$. 

To illustrate the basic idea, let us consider Gaussian distributions $\mathcal{P}_{\Theta}$ as a parametric class, which is closed under convolution, and a Gaussian kernel. For simplicity, we consider the case of scalar variance matrices $\sigma^2 I_d$.  Let $N_d(\mu,\sigma^2 I_d)$ and $f_d(x|\mu,\sigma^2 I_d)$ denote the  $d$-dimensional Gaussian distribution with mean $\mu$ and variance-covariance matrix $\sigma^2 I_d$, and its pdf, respectively. If $P$ and $Q$ are Gaussian distributions $N_d(\mu_P,\sigma^2_P I_d)$ and $N_d(\mu_Q,\sigma^2_Q I_d)$, respectively, and $k$ is given by the pdf $f_d(x-y| 0,\tau^2 I)$, it is easy to see that $m_P(x)=f_d(x| \mu_P,(\sigma_P^2+\tau^2)I_d)$ and $\left\langle m_P,m_Q\right\rangle_\mathcal{H} =f_d(\mu_P| \mu_Q,(\sigma_P^2+\sigma^2_Q+\tau^2)I_d)$. The kernel mean value and inner product are thus reduced to simply evaluating Gaussian pdfs, which are given by a parameter update following a specific rule. 
This type of computation appears in various applications: to list a few, \cite{NIPS2012_0015} proposed a support measure classification by considering kernels $k(P, Q)$ between two input probability measures $P,Q$, including Gaussian models; \cite{TailoringICML2008} and \cite{McCalman2013} considered an approximation of a (target) probability measure $P$ with a Gaussian mixture $P_{\theta}$, via an optimization problem $\hat \theta = \mathrm{argmin}_{\theta} ||m_{P}-m_{P_{\theta}}||_{\mathcal H}^2$.  The parametric expression of (iii) and (iv) is especially useful for the optimization of $\theta$ in the class of distributions.  Other such applications are given in Section \ref{sec:ConnectionML}.  

We generalize this closedness or "conjugacy\footnote{Here, the term "conjugacy" is an analogy of the conjugate prior in the Bayes' theorem, where the prior and posterior have the same pdf form in a probabilistic model. }" of Gaussians with respect to kernel means and explore other cases in CID kernels. We then introduce a {\it conjugate} kernel $k$ to parametric models $\mathcal{P}_{\theta}$ and a {\it convolution trick}, so that (iii) and (iv) have the same density form, i.e., there is some parameter update in the class. If $P, Q$ are delta measures $\delta_{x}, \delta_{y} $, then the convolution trick simplifies to the kernel trick. See Proposition \ref{Prop:CIDkernelmean} for a description.  

While a general perspective is obtained from the convolution semigroup $(\mathbb{I}(\mathbb{R}^d), *)$ of infinitely divisible distributions, the pdfs of $\mathbb{I}(\mathbb{R}^d)$ are not tractable in general.  We then explore smaller convolution sub-semigroups $(\mathcal{P}_{\Theta},*) \subset (\mathbb{I}(\mathbb{R}^d), *)$ having a small number of parameters.  In particular, we focus on the well-known $\alpha$-stable distributions $\mathbb{S}_{\alpha}(\mathbb{R}^d)$ for each $\alpha\in(0,2]$ and generalized hyperbolic (GH) distributions $\mathbb{GH}(\mathbb{R}^d)$, which include Laplace, Cauchy, and Student's $t$ distributions. For each $\alpha \in (0,2]$, the class $\mathbb{S}_{\alpha}(\mathbb{R}^d)$ is closed under convolution. The GH class has various convolutional properties, as given in Proposition \ref{Pro:convolutionRelationsubclassmultivariate}.  As in the Gaussian cases, the computation of (iii) and (iv) is realized by the evaluation of pdfs, i.e., evaluation of conjugate kernels, after a parameter update. 

Unfortunately, these conjugate kernels are not generally tractable. However, we can find some subclasses of tractable conjugate kernels. See Section \ref{sec:ComputationConvolutionTrick} for a discussion on the computation of conjugate kernels. Note that $\alpha$-stable and GH distribution classes have many applications; applications of $\mathbb{S}_{\alpha}(\mathbb{R}^d)$ are listed in \cite{Nolan2013references}, and the GH distributions have been applied, e.g., to mathematical finance with the L\'evy processes \citep{LevyProcessesinFinance_PricingFinancialDerivatives, FinancialModellingwithJumpProcesses, Thevariancegammamodelforsharemarketreturns, Thevariancegammaprocessandoptionpricing, ProcessesofnormalinverseGaussiantype, Apparentscaling, Thefinestructureofassetreturns_anempiricalinvestigation}.  Note also that the Mat\'ern kernel \cite[Section 4.2.1]{GaussianProcessesforMachineLearning}, often used in machine learning, is included in this GH class.

The rest of this paper is organized as follows. In Section \ref{sec:Preliminaries}, we review the notions of kernel means, characteristic kernels, and related matters. In Section \ref{sec:characteristckernelsandIDV}, we show that the CID kernels are characteristic p.d.~kernels on $\mathbb{R}^d$. In addition, we present the closedness property with respect to addition, pointwise product, and convolution. In Section \ref{sec:kernelmeansandIDD}, we introduce the absorbing, conjugate kernel and convolution trick for convolution semigroups of infinitely divisible distributions. Section \ref{sec:ConnectionML} lists some motivating examples of kernel machine algorithms involving kernel means and parametric models. Section \ref{sec:ComputationConvolutionTrick} notes the computation of the pdfs of conjugate kernels to realize the convolution trick.

\section{Preliminaries: Kernel Means and Characteristic Kernels} \label{sec:Preliminaries}
In this section, we review kernel means and characteristic kernels restricted to $\mathbb{R}^d$.

Let $\mathbb{P}_d$ be the set of $d \times d $ p.d. matrices. Let $||x||_{\Sigma}=\sqrt{x^{\top} \Sigma x}$, $x \in \mathbb{R}^d$, and $\Sigma \in \mathbb{P}_d$.
Let $L^{1}(\mathbb{R}^d)$ be the absolutely integrable function space on $\mathbb{R}^d$.
Let $C_b(\mathbb{R}^d)$ be the continuous and bounded function space on $\mathbb{R}^d$.

A symmetric function $k:\mathbb{R}^d \times \mathbb{R}^d \rightarrow \mathbb{R}$ is called a {\it p.d. kernel} on $\mathbb{R}^d$ if, for any $n \in \mathbb{N}$, $x_1,\ldots, x_n \in \mathbb{R}^d$, the matrix $G_{ij}=k(x_i,x_j)$, $i,j \in \{1,\ldots,n\}$ is positive-semidefinite. Throughout this paper, we assume a p.d. kernel $k$ is on $\mathbb{R}^d$.
It is known \citep{Aronszajn1950} that every p.d. kernel $k$ has the unique RKHS $\mathcal{H}$, which is a Hilbert space of functions $f:\mathbb{R}^d \rightarrow \mathbb{R}$, satisfying the following: (i) $k(\cdot, x) \in \mathcal{H}$, $\forall x \in \mathbb{R}^d$; (ii) $\mathrm{Span}\{k(\cdot, x)|x \in \mathbb{R}^d \}$ is dense in $\mathcal{H}$; and (iii) the {\it reproducing property} holds:
\begin{eqnarray*}
f(x) = {\left\langle f, k(\cdot,x) \right\rangle _{\mathcal{H}}}, \hspace{3mm} \forall f \in \mathcal{H}, \hspace{3mm} \forall x \in \mathbb{R}^d   \label{eq:reproducingproperty},
\end{eqnarray*}
where ${\left\langle \cdot ,\cdot \right\rangle _{\mathcal{H}}}$ denotes the inner product of $\mathcal{H}$. The map $\Phi: \mathbb{R}^d \rightarrow \mathcal{H}; x \mapsto k(\cdot, x)$ is called a {\it feature map}.

A p.d. kernel $k$ is called {\it bounded} if $\sup_{x \in \mathbb{R}^d} k(x,x) < \infty$. 
A p.d. kernel $k$ is bounded if and only if every $f \in \mathcal{H}$ is bounded \citep[Lemma 4.23]{SupportVectorMachines2008}.
A p.d. kernel $k$ is called {\it separately continuous} if $k(\cdot, x): \mathbb{R}^d \rightarrow \mathbb{R}$ is continuous for all $x \in \mathbb{R}^d$. A p.d. kernel $k$ is bounded and separately continuous if and only if every $f \in \mathcal{H}$ is a bounded and continuous function, i.e., $\mathcal{H} \subset C_{b}(\mathbb{R}^d)$, \citep[Lemma 4.28]{SupportVectorMachines2008}.
A p.d. kernel $k$ is called {\it continuous} if $k$ is separately continuous and $x \mapsto k(x,x)$, $x \in \mathbb{R}^d$ is continuous \citep[Lemma 4.29]{SupportVectorMachines2008}.
If a p.d. kernel $k$ is continuous, the RKHS $\mathcal{H}$ is separable \citep[Lemma 4.33]{SupportVectorMachines2008}.

A p.d. kernel $k$ is called {\it shift-invariant} if there exists a function $\kappa: \mathbb{R}^d \rightarrow \mathbb{R}$ such that $k(x,y)= \kappa(x-y)$, $x,y \in \mathbb{R}^d$.
The function $\kappa$ is called a {\it p.d. function}. A p.d. function $\kappa$ on $\mathbb{R}^d$ is characterized by the Bochner theorem:
\begin{Theorem} \citep{Bochner1959} \citep[Theorem 6.6]{Wendland2005} \label{Theorem:Bochner}
A continuous function $\kappa : \mathbb{R}^{d} \rightarrow \mathbb{C}$ is positive definite if and only if it is the Fourier transform $\mathcal{F}(\Lambda)$ of a finite nonnegative Borel measure $\Lambda$ on $\mathbb{R}^{d}$:
\begin{eqnarray*}
\kappa (x) =  \int_{{\mathbb{R}^d}}^{} {{e^{\sqrt { - 1} {w ^{\top}}x}}d\Lambda (w )}, \hspace{3mm} x \in \mathbb{R}^{d}.\label{eq:BochnerTheorem}
\end{eqnarray*}
\end{Theorem}
Let $\mathcal{K}_{cb}(\mathbb{R}^d) \subset C_b(\mathbb{R}^d)$ denote the set of continuous bounded p.d. functions.

A p.d. kernel $k$ is called {\it radial} if there exists a function $\tilde \kappa: [0,\infty) \rightarrow \mathbb{R}$ such that $k(x,y)=  \tilde \kappa(|| x-y||)$, $x,y \in \mathbb{R}^d$.
A radial kernel $k$ is given by
\begin{eqnarray}
k(x,y)=\tilde \kappa (||x-y ||) = \int_{[0, \infty)} e^{-t ||x-y||^2}d\nu(t), \hspace{2mm} x, y \in \mathbb{R}^d, \label{eq:radial_integral}
\end{eqnarray}
where $\nu(t)$ is a finite nonnegative Borel measure on the Borel sets $\mathcal{B}([0, \infty))$. A p.d. kernel $k$ is called {\it elliptical} if $k(x,y)= \tilde{\kappa}(|| x-y||_{\Sigma})$, $x,y \in \mathbb{R}^d$, $\Sigma \in \mathbb{P}_d$.

Let $\mathcal{M}_{1}(\mathbb{R}^d)$ be the set of Borel probability measures on $(\mathbb{R}^d, \mathcal{B}(\mathbb{R}^d))$. An RKHS element $m_P \in \mathcal{H}$ with a p.d. kernel $k$ is called a {\it kernel mean} of a probability measure $P \in \mathcal{M}_1(\mathbb{R}^d)$ if there exists an expectation of the feature map:
\begin{eqnarray*}
{m_P} := \mathbb{E}_{X \sim P}[\Phi(X)] ={\mathbb{E}_{X \sim P}}[k( \cdot ,X)] \in \mathcal{H},  \hspace{3mm} P \in \mathcal{M}_1(\mathbb{R}^d). \label{eq:MeanEmbedding}
\end{eqnarray*}
If $k$ is a bounded and continuous p.d. kernel, then the feature map $\Phi:\mathbb{R}^d \rightarrow \mathcal{H}$ is Bochner $P$-integrable for all $P \in \mathcal{M}_1(\mathbb{R}^d)$, since $\mathbb{E}_{X \sim P}[||k(\cdot,X) ||_{\mathcal{H}}] = \mathbb{E}_{X \sim P}[\sqrt{k(X,X)}]<\infty$ for all $P \in \mathcal{M}_1(\mathbb{R}^d)$ \citep[p. 510]{SupportVectorMachines2008}. Throughout this paper, we assume a bounded and continuous p.d. kernel $k$. We write $m_{\mathcal{P}} := \{m_P| P \in \mathcal{P} \subset \mathcal{M}_1(\mathbb{R}^d)  \}$.

As mentioned in the Introduction, there are many applications using $m_P$, since $m_P$ enables us to introduce a similarity and distance between probability measures $P, Q \in \mathcal{M}_1(\mathbb{R}^d)$, via the Hilbert space inner product $\langle m_P, m_Q \rangle_{\mathcal{H}}$ and norm $||m_P- m_Q  ||_{\mathcal{H}}$, respectively, where the reproducing property is also exploited.
In these applications, the characteristic kernel is important to distinguish any probability measures $P, Q \in \mathcal{M}_1(\mathbb{R}^d)$ by their kernel means $m_P, m_Q \in \mathcal{H}$. The following is the definition restricted to $\mathbb{R}^d$:
\begin{Definition}\citep{DBLP:journals/jmlr/FukumizuBJ03}{\cite[Definition 6]{Sriperumbudur_JMLR2010}}
A bounded and continuous p.d. kernel $k:\mathbb{R}^d \times \mathbb{R}^d \rightarrow \mathbb{R}$ is called characteristic on $\mathbb{R}^d$ if the kernel mean map $\mathcal{M}_1(\mathbb{R}^d) \rightarrow \mathcal{H}$; $P \mapsto m_{P}$ is injective, i.e., $m_{P}=m_{Q}$ implies $P=Q$ for any $P,Q \in \mathcal{M}_1(\mathbb{R}^d)$.
\end{Definition}

\cite{Sriperumbudur_JMLR2010} showed a necessary and sufficient condition for a shift-invariant p.d. kernel $k(x,y)= \kappa(x-y)$, $x,y \in \mathbb{R}^d$, $\kappa \in \mathcal{K}_{cb}(\mathbb{R}^d)$, to be characteristic via the Bochner theorem:

\begin{Theorem}{\citep[Theorem 9]{Sriperumbudur_JMLR2010}} \label{Thm:CharacteristicKernelShiftInvariantEucledian}
A shift-invariant p.d. kernel $k$ with $\kappa \in \mathcal{K}_{cb}(\mathbb{R}^d)$ is characteristic if and only if the finite nonnegative measure $\Lambda$ in Theorem \ref{Theorem:Bochner} has the entire support, $\rm{supp}(\Lambda)=\mathbb{R}^{d}$.
\end{Theorem}
Let $\mathcal{K}_{cb}^{ch}(\mathbb{R}^d) \subset \mathcal{K}_{cb}(\mathbb{R}^d)$ denote the set of such characteristic p.d. functions on $\mathbb{R}^d$.

The convolution $f*g$ of two functions $f$ and $g$ is defined by $f*g := \int_{\mathbb{R}^d} f( \cdot -y)g(y)dy$. The convolution $f*Q$ of a function $f$ and a probability measure $Q \in \mathcal{M}_1(\mathbb{R}^d)$ is defined by $f*Q := \int_{\mathbb{R}^d} f(\cdot -y)dQ(y)$. The convolution $P*Q$ of two probability measures $P, Q \in \mathcal{M}_1(\mathbb{R}^d)$ is defined by the probability measure $(P * Q )(B) := \int _{\mathbb{R}^d} P(B-x)dQ(x)$, where $B-x := \{z-x: z \in B \}$, $B \in  \mathcal{B}(\mathbb{R}^d)$.

Given a function $f(x)$, $x \in \mathbb{R}^d$, the function $\tilde f $ denotes $\tilde f(x) = f(-x)$, $x \in \mathbb{R}^d$.
Given a probability measure $P \in \mathcal{M}_1(\mathbb{R}^d)$, a probability measure $\tilde{P} \in \mathcal{M}_1(\mathbb{R}^d)$ is called \textit{dual} if $\tilde{P}(B)=P(-B)$, $B \in \mathcal{B}(\mathbb{R}^{d})$, where $-B:=\{-x:x \in B \}$ \cite[p.8]{Sato1999}.
A probability measure $P$ is symmetric if $P=\tilde{P}$.

We have the following simple equalities:
\begin{Proposition} \label{Prop:convolution_dual}
$\widetilde{f*g} = \tilde f * \tilde g$, $\widetilde{f*P} = \tilde f * \tilde P$, and $\widetilde{P*Q} = \tilde P * \tilde Q$.
\end{Proposition}

Kernel mean $m_P$ and RKHS inner product ${\left\langle {{m_P},{m_Q}} \right\rangle _{\mathcal{H}}}$ have the following convolution representation:

\begin{Lemma} \label{lm:kernelmeanprobabilitydensityfunction}
Let $k$ be a shift-invariant p.d. kernel with $\kappa \in C_b(\mathbb{R}^d)$. Then, we have the following:
\begin{enumerate}
\item Kernel mean $m_{P}$ is given by the convolution
\begin{eqnarray*}
 m_P = \kappa  * P \in \mathcal{H} \subset C_{b}(\mathbb{R}^d),  \hspace{3mm} P \in \mathcal{M}_1(\mathbb{R}^{d}).
\end{eqnarray*}
\item The RKHS inner product ${\left\langle {{m_P},{m_Q}} \right\rangle _{\mathcal{H}}}$ is given by the convolution
\begin{eqnarray*}
{\left\langle {{m_P},{m_Q}} \right\rangle _{\mathcal{H}}} =(\kappa * \tilde P *Q)(0) =(\kappa * P *\tilde Q)(0), \hspace{3mm} P,Q \in \mathcal{M}_1(\mathbb{R}^{d}),
\end{eqnarray*}
where $\tilde P$ and $\tilde Q$ are the dual of $P$ and $Q$, respectively.
\end{enumerate}
\end{Lemma}
\begin{proof}
1. Kernel mean $m_P$ has the following convolution representation:
\begin{eqnarray*}
m _P = \int_{\mathbb{R}^{d}}{k(x,\cdot)dP(x)}  = \int_{\mathbb{R}^{d}}{\kappa (\cdot -x)dP(x)}  =  \kappa  * P, \hspace{3mm}  P \in \mathcal{M}_1(\mathbb{R}^{d}).
\end{eqnarray*}
Kernel mean $m_P \in \mathcal{H} \subset C_{b}(\mathbb{R}^d)$ exists for all $P \in \mathcal{M}_1(\mathbb{R}^d)$ because, for $\kappa \in C_b(\mathbb{R}^d)$, the feature map $\Phi: x \mapsto k(x, \cdot)$ is Bochner $P$-integrable for all $P \in \mathcal{M}_1(\mathbb{R}^d)$, as given in the definition of $m_P$.

2. RKHS inner product ${\left\langle {{m_P},{m_Q}} \right\rangle _{\mathcal{H}}}$ has the following convolution representation:
\begin{eqnarray*}
{\left\langle {{m_P},{m_Q}} \right\rangle _{\mathcal{H}}}  =  \int_{\mathbb{R}^{d}} {{m_P}(y)dQ(y)}  = \int_{\mathbb{R}^{d}} {{{\tilde m}_P}( - y)dQ(y)}  = ({{\tilde m}_P}*Q)(0)=(\kappa * \tilde P *Q)(0),
\end{eqnarray*}
where we have used Proposition \ref{Prop:convolution_dual} and $\tilde \kappa = \kappa$ in the last equality. Since ${\left\langle {{m_P},{m_Q}} \right\rangle _{\mathcal{H}}}$ is symmetric with respect to $P$ and $Q$, then $(\kappa * \tilde P *Q)(0) =(\kappa * P *\tilde Q)(0)$. This is also obtained by $({\kappa * \tilde P *Q})(0) = (\widetilde{\kappa * \tilde P *Q})(0) =(\kappa * P *\tilde Q)(0)$.
\end{proof}

\vspace{2mm}
In this paper, we simply consider that $\kappa$ is a pdf of a probability distribution.\footnote{In machine learning, normalized kernels $\bar k(x,y):=\frac{k(x,y)}{\sqrt{k(x,x)}\sqrt{k(y,y)}}$ are often used (e.g., Gaussian kernels $\bar k(x,y) := \exp(-\frac{||x-y ||^2}{2 \gamma^2})$) \citep[Lemma 4.55]{SupportVectorMachines2008}. However, we consider here pdf kernels (e.g., Gaussian kernels $k(x,y) := \frac{1}{\sqrt{(2\pi \gamma^2)^d} } \exp(-\frac{||x-y ||^2}{2 \gamma^2})$) for the closedness of the pdfs of $P$ and $m_P$. A scalar multiplication ($c>0$) changes as follows: $\bar m_P :=\mathbb{E}_{X \sim P}[\bar k(\cdot, X)] = c \mathbb{E}_{X \sim P}[k(\cdot, X)] =cm_P$ and ${\left\langle {{\bar m_P},{\bar m_Q}} \right\rangle _{\bar{\mathcal{H}}}} = c {\left\langle {{m_P},{m_Q}} \right\rangle _{\mathcal{H}}} $, where ${\left\langle {f,g} \right\rangle _{\bar{\mathcal{H}}}} = \frac{1}{c} {\left\langle f, g\right\rangle _{\mathcal{H}}} $, $\forall f,g \in \mathcal{H}, \bar{\mathcal{H}}$ \cite[p.37]{BerlinetBook2004}.  }
Then, Lemma \ref{lm:kernelmeanprobabilitydensityfunction} motivates us to explore the set of probability distributions $\mathcal{P}_{\Theta} \subset \mathcal{M}_{1}(\mathbb{R}^d)$ that is closed under convolution, i.e., convolution semigroup $(\mathcal{P}_{\Theta}, *)$.

\section{Characteristic Kernels and Infinitely Divisible Distributions} \label{sec:characteristckernelsandIDV}
In this section, we introduce CID kernels, which are defined by infinitely divisible distributions, and show that they are characteristic (Section \ref{sec:CIDkernels}). In addition, we examine some closure properties of CID kernels with respect to addition, pointwise product, and convolution (Section \ref{sec:closurepropertyCID}).
\subsection{Convolutionally Infinitely Divisible Kernels} \label{sec:CIDkernels}
We review the infinite divisibility of a probability measure \citep{Sato1999, InfinitelyDivisibleRealLine2004, Applebaum2009}.
\begin{Definition}{\cite[Definition 7.1, p. 31]{Sato1999}}
A probability measure $P \in \mathcal{M}_1(\mathbb{R}^d)$ is called infinitely divisible if, for any integer $n \in \mathbb{N}$, there exists a probability measure $P_{n} \in \mathcal{M}_1(\mathbb{R}^d)$ such that $P=P_{n}^{*n}$.
\end{Definition}
The support of every infinitely divisible distribution $P$ is unbounded except for delta measures $\{ \delta_x(\cdot) |x \in \mathbb{R}^d\}$ \cite[Examples 7.2, p. 31]{Sato1999}. 
Let $\mathbb{I}(\mathbb{R}^{d})$ denote the set of infinitely divisible distributions on $\mathbb{R}^d$. $\mathbb{I}(\mathbb{R}^{d})$ is closed under convolution.
Every infinitely divisible distribution $P \in \mathbb{I}(\mathbb{R}^{d})$ has the following unique {\it L\'evy--Khintchine representation} for the characteristic function. 
Let $x \wedge y = \min \{x,y\}$, $x,y \in \mathbb{R}$. Let $1_{B}$ denote the indicator function on $\mathbb{R}^{d}$ with $B \subset \mathbb{R}^{d}$.
\begin{Theorem}{\cite[Theorem 8.1, p. 37]{Sato1999}}
The characteristic function $\hat P (w )$ of an infinitely divisible distribution $P \in \mathbb{I}(\mathbb{R}^{d})$ has the following unique representation:
\begin{eqnarray}
\hat P(w ) = \exp \left( {i w^{\top} \gamma  - \frac{1}{2} w^{\top} A w   + \int_{{\mathbb{R}^d}}^{} {\left( {{e^{i w^{\top} x }} - 1 - i w^{\top} x {{1}_{\{ |x| \le 1\} }}(x)} \right)\nu(dx)} } \right), \hspace{1mm} w \in  \mathbb{R}^{d}, \label{eq:chfuncInfinitelydivisible}
\end{eqnarray}
where $\gamma \in \mathbb{R}^{d}$, $A \in \mathbb{R}^{d \times d}$, is a symmetric nonnegative-definite matrix and $\nu$ is a measure on $\mathbb{R}^{d}$ satisfying
\begin{eqnarray}
\nu(\{\mathbf{0}\})=0 \hspace{3mm} and \hspace{3mm} \int_{{\mathbb{R}^d}}^{} {(|x{|^2} \wedge 1)\nu (dx)}  < \infty .  \label{eq:conditionLevymeasure}
\end{eqnarray}
Conversely, for any $\gamma \in \mathbb{R}^{d}$, symmetric nonnegative-definite matrix $A \in \mathbb{R}^{d \times d}$, and measure $\nu$ satisfying  (\ref{eq:conditionLevymeasure}), there exists an infinitely divisible distribution $P \in \mathbb{I}(\mathbb{R}^{d})$.
\end{Theorem}
$(A,\nu,\gamma)$ is called the {\it generating triplet} of $P \in \mathbb{I}(\mathbb{R}^{d})$. $A$ is called the covariance matrix of the Gaussian factor of $P \in \mathbb{I}(\mathbb{R}^{d})$, and $\nu$ is called the {\it L\'evy measure} of $P \in \mathbb{I}(\mathbb{R}^{d})$. Gaussians correspond to the generating triplet $(A,0,\gamma)$. $\alpha$-Stable distributions, including Cauchy distributions, correspond to generating triplet $(0,\nu,\gamma)$, where $\nu$ is the corresponding nonzero L\'evy measure. The L\'evy measure of the $\alpha$-stable distributions is shown in Appendix \ref{sec:alpha-stableDist}.

An infinitely divisible distribution $P \in \mathbb{I}(\mathbb{R}^{d})$ is symmetric if and only if $(A,\nu,\gamma)=(A,\nu_s,0)$, where $\nu_s$ is a symmetric L\'evy measure\footnote{A symmetric L\'evy measure is a L\'evy measure such that $\nu_s (B) = \nu_s (-B)$ for $\forall B \in \mathcal{B}(\mathbb{R}^d)$.}  \cite[p.114]{Sato1999}. Let $\mathbb{IS}(\mathbb{R}^{d})$ denote the set of symmetric and infinitely divisible distributions on $\mathbb{R}^d$. $\mathbb{IS}(\mathbb{R}^{d})$ is closed under convolution.
Let $\mathcal{K}_{cb}^{id}(\mathbb{R}^d) (\subset C_b(\mathbb{R}^d) \cap L^1(\mathbb{R}^d))$ denote the set of continuous and bounded pdfs\footnote{A necessary and sufficient condition for $P \in \mathbb{IS}(\mathbb{R}^d)$ to have a pdf is not known \cite[p.177]{Sato1999}. If the Gaussian factor $A \in \mathbb{R}^{d \times d}$ is full rank, then $P \in \mathbb{I}(\mathbb{R}^d)$ has the pdf. If $A=0$, see some sufficient conditions \cite[Theorem 27.7, 27.10]{Sato1999}. Every nondegenerate self-decomposable distribution on $\mathbb{R}^d$ has the pdf \cite[Theorem 27.13]{Sato1999}.} of symmetric infinitely divisible distributions $\mathbb{IS}(\mathbb{R}^{d})$:
\begin{eqnarray*}
\mathcal{K}_{cb}^{id}(\mathbb{R}^d) := \{ \Xi(P_s) \in C_b(\mathbb{R}^d) | P_s \in \mathbb{IS}(\mathbb{R}^{d}) \},
\end{eqnarray*}
where $\Xi: \mathcal{M}_1(\mathbb{R}^d) \rightarrow L^1(\mathbb{R}^d)$ is a function that maps a probability measure $P$ to its pdf $f$ if it exists.

The infinitely divisible pdf $\kappa \in \mathcal{K}_{cb}^{id}(\mathbb{R}^d)$ can be used for a characteristic kernel as follows.

\begin{Theorem} \label{thm:characteristic_convolutionIDkernel}
The function $k(x,y)=\kappa(x-y)$, $x,y \in \mathbb{R}^{d}$, $ \kappa \in \mathcal{K}_{cb}^{id}(\mathbb{R}^d)$ is a p.d. and characteristic kernel, i.e., $\mathcal{K}_{cb}^{id}(\mathbb{R}^d) \subset \mathcal{K}_{cb}^{ch}(\mathbb{R}^d)$.
\end{Theorem}
\begin{proof}
A probability measure $P$ on $\mathbb{R}^{d}$ is symmetric if and only if the characteristic function $\hat P(w)$, $w \in \mathbb{R}^{d}$ is real valued \cite[p.67]{Sato1999}. If $P$ is symmetric and infinitely divisible, $\hat P(w)>0$ for every $w \in \mathbb{R}^{d}$ from the L\'evy--Khintchine formula (\ref{eq:chfuncInfinitelydivisible}).
Since $\hat P(w)$ is positive and has the entire support, $ {\rm supp}(\hat P(w))=\mathbb{R}^d$, then $k$ is a p.d. and characteristic kernel from Theorem \ref{Thm:CharacteristicKernelShiftInvariantEucledian}.
\end{proof}
We call a p.d. kernel $k$ in Theorem \ref{thm:characteristic_convolutionIDkernel} a {\it convolutionally infinitely divisible} (CID) kernel\footnote{The term "infinite divisibility" of a p.d. kernel is used in the pointwise product sense \cite[Definition 2.6, p. 76]{BergChristensen1984}, i.e., a p.d. kernel $k:\mathcal{X} \times \mathcal{X} \rightarrow \mathbb{C}$ on a nonempty set $\mathcal{X}$ is called {\it infinitely divisible} if, for every $n \in \mathbb{N}$, there exists a p.d. kernel $k_{n}:\mathcal{X} \times \mathcal{X} \rightarrow \mathbb{C}$ such that $k=(k_{n})^{n}$. The CID kernel considered here is the convolution sense $\kappa = (\kappa_n)^{*n}$.
}.
CID kernels include the following examples:
\begin{Example}[CID p.d. kernels] \label{ex:symmetricinfinitedivisibledistribution}
CID kernels include Gaussian kernels, Laplace kernels, Cauchy kernels, $\alpha$-stable kernels for each $\alpha \in (0,2]$ ($\alpha=2$ corresponds to Gaussian kernels; $\alpha=1$ corresponds to Cauchy kernels), sub-Gaussian $\alpha$-stable kernels, Student's $t$ kernels \citep{SelfDecomposabilityStudentT1976}, GH kernels, normalized inverse Gaussian (NIG) kernels, variance gamma (VG) kernels (Mat\'ern kernel is a special case of this), tempered $\alpha$-stable (T$\alpha$S) kernels \citep{VolatilityClustering2011, Rosinski2007, Bianchi2010}, etc.
\end{Example}

\subsection{Closure Property} \label{sec:closurepropertyCID}
In this subsection, we note some closure properties of CID and characteristic kernels with respect to addition, pointwise product, and convolution.
The closure property is used, e.g., to generate a new CID and characteristic kernel. Example \ref{eq:LaplaceVG_infinitelydivisible} shows such an example.

It is known that the set of continuous and bounded p.d. kernels $\mathcal{K}_{cb}(\mathbb{R}^d)$ is closed under addition and pointwise product as follows \citep[p. 114]{SupportVectorMachines2008}:
 \begin{Proposition} \label{Prop:ClosedPropertyPDKernel}
If $\kappa_1, \kappa_2 \in \mathcal{K}_{cb}(\mathbb{R}^d)$, then $\kappa_1 + \kappa_2 \in \mathcal{K}_{cb}(\mathbb{R}^d)$ and $\kappa_1\kappa_2 \in \mathcal{K}_{cb}(\mathbb{R}^d)$.
\end{Proposition}

Similarly, the set of characteristic kernels $\mathcal{K}_{cb}^{ch}(\mathbb{R}^d)$ is closed under addition and pointwise product as follows \cite[Corollary 11]{Sriperumbudur_JMLR2010}:
\begin{Proposition} \label{Prop:ClosedPropertyCharacteristicKernel}
If $\kappa \in \mathcal{K}_{cb}^{ch}(\mathbb{R}^d)$, $\kappa_1, \kappa_2 \in \mathcal{K}_{cb}(\mathbb{R}^d)$, and $\kappa_2 \neq 0$, then $\kappa + \kappa_1, \kappa \kappa_2 \in \mathcal{K}_{cb}^{ch}(\mathbb{R}^d)$.
\end{Proposition}
The set of CID kernels $\mathcal{K}_{cb}^{id}(\mathbb{R}^d)$ is closed under convolution but not closed under addition or pointwise product.
\begin{Proposition} \label{Prop:CIDcloseness}
Let $\kappa_1, \kappa_2 \in \mathcal{K}_{cb}^{id}(\mathbb{R}^d)$. Then, we have the following:
\begin{enumerate}
\item Convolution $\kappa_1 * \kappa_2 \in \mathcal{K}_{cb}^{id}(\mathbb{R}^d)$.
\item Addition $\kappa_1 + \kappa_2$ and product $\kappa_1 \kappa_2$ do not necessarily belong to $\mathcal{K}_{cb}^{id}(\mathbb{R}^d)$, although they are characteristic, $\kappa_1 + \kappa_2, \kappa_1\kappa_2 \in \mathcal{K}_{cb}^{ch}(\mathbb{R}^d)$.
\end{enumerate}
\end{Proposition}
\begin{proof}
1. Let $\kappa_1 = \Xi(P_1)$ and $\kappa_2 = \Xi(P_2)$. Then, $\kappa_1 * \kappa_2 = \Xi(P_1 * P_2)$. If $P_1, P_2 \in \mathbb{IS}(\mathbb{R}^d)$ are absolutely continuous and symmetric infinitely divisible measures, so is $P_1 * P_2 \in \mathbb{IS}(\mathbb{R}^d)$. 

2. A mixture of two infinitely divisible distributions is not necessarily infinitely divisible. A product of two infinitely divisible distributions is not necessarily infinitely divisible. The counter-examples are as follows. Let $\kappa_1(x) = e^{-|x|}$ and $\kappa_2(x) = e^{-x^2}$, $x \in \mathbb{R}$, be p.d. functions of Laplace and Gaussian kernels, respectively. Then, the product $k(x) \propto e^{-|x|}e^{-x^2}$ is not infinitely divisible \citep[Example 11.13]{InfinitelyDivisibleRealLine2004}, although it is characteristic (Proposition \ref{Prop:ClosedPropertyCharacteristicKernel}). Let $\kappa_1(x) = \frac{1}{4 \sqrt{\pi}} e^{-\frac{1}{4}x^2}$ and $\kappa_2(x) = \frac{1}{4 \sqrt{2 \pi}} e^{-\frac{1}{8}x^2}$, $x \in \mathbb{R}$, be Gaussian kernels; then, the addition $\kappa_1+ \kappa_2$ is not infinitely divisible \citep[Example 11.15]{InfinitelyDivisibleRealLine2004}, although it is characteristic (Proposition \ref{Prop:ClosedPropertyCharacteristicKernel}). Many examples can be found in \cite{InfinitelyDivisibleRealLine2004}.
\end{proof}
As given in Proposition \ref{Prop:CIDcloseness}, the infinite divisibility is not closed under mixing in general, although some special mixing cases preserve it \cite[Chapter 7]{InfinitelyDivisibleRealLine2004}. The {\it normal mean-variance mixture} with an infinitely divisible mixing distribution, given in Lemma \ref{Lemma:PropertyNormalMeanVarianceMixture}, is one of them.

New CID kernels and characteristic kernels may be generated by using these closure properties. If $\kappa = \mathcal{F} (\hat \kappa)$ is an infinitely divisible pdf with the characteristic function $\hat \kappa$, then symmetrization $\kappa^*:=\kappa* \tilde {\kappa} = \mathcal{F} (|\hat \kappa|^2)$ and positive powers $(\kappa^*)^{*\lambda} =\mathcal{F} (|\hat \kappa|^{2\lambda})$ $(\lambda>0)$ are also infinitely divisible pdfs.
The following example shows that the Laplace and symmetric Gamma kernels are CID kernels generated from an exponential distribution.

\begin{Example}{\cite[Example 2.9]{InfinitelyDivisibleRealLine2004}} \label{eq:LaplaceVG_infinitelydivisible}
An exponential distribution $P$ with the pdf $\kappa(x)=\alpha \exp (- \alpha x)1_{[0,\infty)}(x)$, $\alpha>0$ is infinitely divisible. The dual is $\tilde{\kappa}(x)=\alpha \exp (\alpha x)1_{(-\infty,0)}(x)$.
\begin{enumerate}
\item The symmetrization $\kappa^* = \kappa *\tilde{ \kappa} $ has the characteristic function ${{\hat \kappa}^*}(w)=\hat \kappa (w) \hat {\tilde{ \kappa}}(w) = \frac{\alpha }{{\alpha  - iw  }} \cdot \frac{\alpha }{{\alpha  + iw }} = \frac{{{\alpha ^2}}}{{{\alpha ^2} + {w ^2}}}$. This is a Laplace pdf $\kappa^{*}(x)= \frac{\alpha}{2}\exp (-\alpha |x|)$.
\item Positive powers $(\kappa^*)^{*\lambda}$ $(\lambda>0)$ have the characteristic functions $({{\hat \kappa}^*})^{\lambda}(w) = ( \frac{{{\alpha ^2}}}{{{\alpha ^2} + {w ^2}}} )^\lambda$. If $\lambda=1$, the pdf is the above Laplace case. If $\lambda=2$, the pdf is given by $(\kappa^{*})^{*2}(x)= \frac{\alpha}{4}(1+\alpha |x| )\exp (-\alpha |x|)$. For general $\lambda>0$, the pdf is given by
\begin{eqnarray*}
f(x)=\frac{\alpha ^{2 \lambda}}{\sqrt{\pi}(2 \alpha)^{\lambda - \frac{1}{2}}\Gamma(\lambda)} |x-\mu|^{\lambda -\frac{1}{2}} K_{\lambda - \frac{1}{2}}(\alpha |x-\mu|), \hspace{2mm} x \in \mathbb{R}
\end{eqnarray*}
where $\Gamma(\lambda)$ is the Gamma function and $K_{\lambda}(x)$ is the modified Bessel function of the third kind with index $\lambda$. This is the pdf of the zero-skewed VG distribution $VG_1(\lambda, \alpha, \beta=0,\mu, 1)$ on $\mathbb{R}$, as given in Section \ref{sec:GHdistributionsbody}.
\end{enumerate}
The additions $(\kappa^{*})^{*\lambda} + \tilde \kappa$, $\tilde \kappa \in \mathcal{K}_{cb}^{}(\mathbb{R}^d)$, and products $(k^{*})^{*\lambda}\tilde \kappa$, $\tilde \kappa \in \mathcal{K}_{cb}^{ch}(\mathbb{R}^d)$, are characteristic kernels based on the closure properties.
\end{Example}

\section{Kernel Means and Infinitely Divisible Distributions} \label{sec:kernelmeansandIDD}
In this section, we examine the kernel means of a parametric class of distributions $\mathcal{P}_{\Theta} \subset \mathbb{I}(\mathbb{R}^d)$. As mentioned in the Introduction, we wish to compute (iii) kernel mean values $m_{P}(x)$, $x \in \mathbb{R}^d$ and (iv) RKHS inner products ${\left\langle {{m_P},{m_Q}} \right\rangle _{\mathcal{H}}}$ for parametric models $P, Q \in \mathcal{P}_{\Theta}$. These form a basic computation for establishing kernel machine algorithms combining kernel means and parametric models. 
In Section \ref{sec:absorbingconjugate}, we introduce absorbing, conjugate kernels, and convolution trick in the set of infinitely divisible distributions $\mathbb{I}(\mathbb{R}^d)$. In Sections \ref{sec:stabledistributionbody} and \ref{sec:GHdistributionsbody},
we focus on well-known subclasses of $\alpha$-stable distributions and GH distributions, which include Laplace, Cauchy, and Student's $t$ distributions.

\subsection{Absorbing, Conjugate Kernels, and Convolution Trick} \label{sec:absorbingconjugate}
We begin by introducing the notion of {\it absorbing} and {\it conjugate} p.d kernels to particular sets of parametric models $\mathcal{P}_{\Theta}$ as follows:
\begin{Proposition}[absorbing \& conjugate kernel] \label{Def:AbsorbingKernel}
Let $\mathcal{P}_{\Theta}, \mathcal{Q}_{\Theta'} \subset \mathcal{M}_1(\mathbb{R}^d)$ be two sets of parametric models such that $\mathcal{P}_{\Theta} * \mathcal{Q}_{\Theta'} \subseteq \mathcal{P}_{\Theta}$, where $\Theta$ and $\Theta'$ are finite or infinite index sets. Denote by $\Xi(\mathcal{P}_{\Theta})$ and $\Xi(\mathcal{Q}_{\Theta'})$ the sets of pdfs. Let $\kappa \in \mathcal{K}_{cb}(\mathbb{R}^d)$ be a shift-invariant p.d. kernel. We have the following statements:

\begin{enumerate}
\item If $\kappa \in \Xi(\mathcal{P}_{\Theta})$, then $m_{\mathcal{Q}_{\Theta'}} \subset \Xi(\mathcal{P}_{\Theta})$ holds. RKHS inner products ${\left\langle {{m_P},{m_Q}} \right\rangle _{\mathcal{H}}}$, $P, Q \in \mathcal{Q}_{\Theta'}$ are values of pdfs in $\Xi(\mathcal{P}_{\Theta})$.
\item If $\kappa \in \Xi(\mathcal{Q}_{\Theta'})$, then $m_{\mathcal{P}_{\Theta}} \subset \Xi(\mathcal{P}_{\Theta})$ holds. RKHS inner products ${\left\langle {{m_P},{m_Q}} \right\rangle _{\mathcal{H}}}$, $P, Q \in \mathcal{P}_{\Theta}$ are not necessarily values of pdfs in $\Xi(\mathcal{P}_{\Theta})$.
\end{enumerate}
\end{Proposition}
\begin{proof}
These statements are straightforward from Lemma \ref{lm:kernelmeanprobabilitydensityfunction} and assumptions.
\end{proof}
Statements $1$ and $2$ indicate an {\it absorbing property} of $k$ with respect to parametric models. 
If $\mathcal{P}_{\Theta} = \mathcal{Q}_{\Theta'}$ in Proposition \ref{Def:AbsorbingKernel}, we call $k$ (and, hence, its RKHS $\mathcal{H}$)  a {\it conjugate} to $\mathcal{P}_{\Theta}$. A general perspective may be given by the CID kernels, where these kernels are conjugate to $\mathbb{I}(\mathbb{R}^{d})$ as follows: 
\begin{Proposition} \label{Prop:CIDkernelmean}
Let $k_{A,\nu_s}(x, y)=\kappa_{A,\nu_s}(x-y)$, $x,y \in \mathbb{R}^d$ be a CID kernel, where $\kappa_{A,\nu_s} \in \mathcal{K}_{cb}^{id}(\mathbb{R}^d)$ has a generating triplet $(A, \nu_s, 0)$, and let $\mathcal{H}_{A,\nu_{s}}$ be the RKHS given by $\kappa_{A,\nu_s}$. Let $P, Q \in \mathbb{I}(\mathbb{R}^{d})$ be infinitely divisible distributions with the generating triplets $(A_P, \nu_P, \gamma_P)$ and $(A_Q, \nu_Q, \gamma_Q)$, respectively. Then, we have the following:
\begin{enumerate}
\item Kernel mean $m_P$ is given by an infinitely divisible pdf:
\begin{eqnarray*}
m_{P}(\cdot)&=& f (\cdot;A+A_P,\nu_{s}+\nu_P,\gamma_P), \hspace{3mm} f \in \Xi(\mathbb{I}(\mathbb{R}^d)) \\
 &=& k_{A+A_P,\nu_{s}+{\nu}_{P}}(\gamma_{P}, \cdot)
. \label{eq:kernelmeanInfiniteDivisible}
\end{eqnarray*}
\item The RKHS inner product ${\left\langle {{m_P},{m_Q}} \right\rangle _{\mathcal{H}_{A, \nu_s}}} $ is given by
\begin{eqnarray*}
{\left\langle {{m_P},{m_Q}} \right\rangle _{\mathcal{H}_{A, \nu_s}}} &=& f(0 ;A+A_P+A_Q,\nu_{s}+\tilde{\nu}_{P}+\nu_{Q},\gamma_{Q}-\gamma_{P}) \\
&=& f(0 ;A+A_P+A_Q,\nu_{s}+{\nu}_{P}+ \tilde{\nu}_{Q},\gamma_{P}-\gamma_{Q}), \\
&=& k_{A+A_P+A_Q,\nu_{s}+{\nu}_{P}+ \tilde{\nu}_{Q}}(\gamma_{P}, \gamma_{Q}), \\
\end{eqnarray*}
where $\tilde{\nu}_{P}$ (respectively, $\tilde{\nu}_{Q}$) is the dual of the L\'evy measure $\nu_{P}$ (respectively, $\nu_{Q}$).
\end{enumerate}
\end{Proposition}
Proposition \ref{Prop:CIDkernelmean} indicates a general {\it convolution trick}. 
The computation of ${\left\langle {{m_P},{m_Q}} \right\rangle _{\mathcal{H}_{A, \nu_s}}} $ is reduced to the computation of the same kernel $k_{A+A_P+A_Q,\nu_{s}+{\nu}_{P}+ \tilde{\nu}_{Q}}$ with the updated parameters of the generating triplets. If $Q$ is a delta measure $\delta_y$ (i.e., $A_Q=0$, $ \nu_Q =0$, $\gamma_Q=y$), then statement 2 is specialized to statement 1.
If $P, Q$ are both delta measures $\delta_x$, $\delta_y$ (i.e., $A_P=A_Q=0$, $\nu_P = \nu_Q =0$, $\gamma_P = x$, $\gamma_Q=y$), then statement 2 is specialized to the kernel trick ${\left\langle {k_{A, \nu_s}(\cdot,x), k_{A, \nu_s}(\cdot,y)} \right\rangle _{\mathcal{H}_{A, \nu_s}}}= k_{A,\nu_{s}}(x, y)$. 
If $P, Q$ and $k$ are all Gaussians (i.e., $\nu_P = \nu_Q = \nu_s=0$), then statement 2 results in the computation of the same Gaussian kernel with increased variance $A+A_P+A_Q$, where the computation of Gaussian pdfs is tractable.

Although Proposition \ref{Prop:CIDkernelmean} gives us a theory that kernel means $m_{\mathcal{P}}$ and RKHS inner products $\langle {{m_P},{m_Q}} \rangle$ are expressed with generating triplets $(A,\nu, \gamma)$, the computation of the general infinitely divisible pdfs may be intractable. We then systematically examine smaller subsemigroups of parametric models $(\mathcal{P}_{\Theta}, *)  \subset (\mathbb{I}(\mathbb{R}^{d}),*)$ such that the computation of pdfs may be possible. 
We specifically examine well-known parametric classes of $\alpha$-stable distributions and GH distributions on $\mathbb{R}^d$ in Sections \ref{sec:stabledistributionbody} and \ref{sec:GHdistributionsbody}, respectively.

\subsection{$\alpha$-stable distributions} \label{sec:stabledistributionbody}
$\alpha$-Stable distributions $\mathbb{S}_{\alpha}(\mathbb{R}^{d})$, $\alpha \in (0,2]$, on $\mathbb{R}^d$ are a well-known convolution subsemigroup of infinitely divisible distributions \citep{Zolotarev1986, Samorodnitsky1994}.

$\alpha=2$ implies Gaussian distributions $\mathbb{S}_2(\mathbb{R}^{d})=\mathbb{G}(\mathbb{R}^{d})$, which are closed under convolution; if $P$ and $Q$ are $N(\mu_P, R_P)$ and $N(\mu_Q, R_Q)$ with mean vectors $\mu_P, \mu_Q$ and covariance matrices $R_P$, $R_Q$, respectively, then convolution $P*Q$ is $N(\mu_P+\mu_Q, R_P+R_Q)$.

For $\alpha \in (0,2)$, $\alpha$-stable distributions are heavy tailed, where there are many applications, as listed in \cite{Nolan2013references}. 
For each $\alpha \in (0,2)$, a one-dimensional $\alpha$-stable distribution $S_{\alpha}(\sigma, \beta, \mu)$ is specified by a scale parameter $\sigma > 0$, a skewness parameter $\beta \in [-1,1]$, and a location parameter $\mu \in \mathbb{R}$. For each $\alpha \in (0,2)$, the set $\mathbb{S}_{\alpha}(\mathbb{R})$ is closed under convolution; if $P$ and $Q$ are two stable laws $S_{\alpha}(\sigma_P, \beta_P, \mu_P)$ and $S_{\alpha}(\sigma_Q, \beta_Q, \mu_Q)$, respectively, then $P*Q$ is $S_{\alpha}(\sigma, \beta, \mu)=S_{\alpha}((\sigma_P^\alpha+\sigma_Q^\alpha)^{1/\alpha}, \frac{\beta_P \sigma_P^\alpha+\beta_Q \sigma_Q^\alpha}{\sigma_P^\alpha+\sigma_Q^\alpha}, \mu_P +\mu_Q)$ \citep[Property1.2.1]{Samorodnitsky1994}. See Appendix \ref{sec:univariatestabledistributionkernel} for more details.

For each $\alpha \in (0,2)$, a $d$-dimensional $\alpha$-stable distribution $S_{\alpha}(\mu, \Gamma)$ is specified by a location parameter $\mu \in \mathbb{R}^d$ and a spectral measure $\Gamma$ on the unit sphere $S_{d-1} := \{s \in \mathbb{R}^d: ||s||=1\}$ \citep[Theorem 2.3.1, p.65]{Samorodnitsky1994}. For each $\alpha \in (0,2)$, the set $\mathbb{S}_{\alpha}(\mathbb{R}^{d})$ is closed under convolution; if $P$ and $Q$ are two stable laws $S_{\alpha}(\mu_P, \Gamma_P)$ and $S_{\alpha}(\mu_Q, \Gamma_Q)$, respectively, then $P*Q$ is $S_{\alpha}(\mu_P+\mu_Q, \Gamma_P +\Gamma_Q)$. See Appendix \ref{sec:multivariatestabledistributionkernel} for more details. 
$\alpha$-Stable pdfs on $\mathbb{R}^d$ are intractable in general. 

Sub-Gaussian $\alpha$-stable distributions (equivalently, elliptically contoured $\alpha$-stable distributions) $\mathbb{SG}_{\alpha}(\mathbb{R}^{d})$ are a well-known subclass of $\mathbb{S}_{\alpha}(\mathbb{R}^d)$ \citep{Samorodnitsky1994, RePEc:spr:compst:v:28:y:2013:i:5:p:2067-2089}.
For each $\alpha \in (0,2)$, a sub-Gaussian $\alpha$-stable distribution is specified by a location parameter $\mu \in \mathbb{R}^d$ and a p.d. matrix $R \in \mathbb{R}^{d \times d}$ \citep[Theorem 2.5.2, p.78]{Samorodnitsky1994}. See Appendix \ref{sec:subGaussclass} for more details. Sub-Gaussian $1$-stable distributions imply $d$-dimensional Cauchy distributions $\mathbb{CAU}(\mathbb{R}^d)$ \citep[Example 2.5.3, p.79]{Samorodnitsky1994}. If $d=1$, for each $\alpha \in (0,2)$, sub-Gaussians $\mathbb{SG}_{\alpha}(\mathbb{R})$ are closed under convolution. If $d>1$, for each $\alpha \in (0,2)$, sub-Gaussians $\mathbb{SG}_{\alpha}(\mathbb{R}^{d})$ are not closed under convolution. Let us decompose $\mathbb{SG}_{\alpha}(\mathbb{R}^{d})$ into an equivalent class $\mathbb{SG}_{\alpha}(\mathbb{R}^{d})= \bigcup_{R} \mathbb{SG}_{\alpha}(\mathbb{R}^{d})[R]$, where
\begin{eqnarray*}
\mathbb{SG}_{\alpha}(\mathbb{R}^{d})[R] := \{P \in \mathbb{SG}_{\alpha}(\mathbb{R}^{d}) \mid P= SG_{\alpha}(\mu, c R), \mu \in \mathbb{R}^d, \hspace{1mm} c >0 \}.
\end{eqnarray*}
For each $\alpha \in (0,2)$ and a p.d. matrix $R \in \mathbb{P}^{d}$, the set $\mathbb{SG}_{\alpha}(\mathbb{R}^{d})[R]$ is closed under convolution; if $P$ and $Q$ are $SG_{\alpha}(\mu_P, c_PR)$ and $SG_{\alpha}(\mu_Q, c_Q R)$, respectively, then $P*Q$ is $SG_{\alpha}(\mu_P +\mu_Q, (c_P^{\frac{\alpha}{2}}+c_Q^{\frac{\alpha}{2}})^{\frac{2}{\alpha}}R)$. Note that when $\alpha=2$, the whole set $\mathbb{SG}_{2}(\mathbb{R}^{d})$ is closed.

These convolution properties of $\alpha$-stable distributions lead to the following conjugate pairs of $\alpha$-stable kernels $k$ and $\alpha$-stable distributions $\mathcal{P}_{\Theta}$.
\begin{Example} \label{ex:conjugate_alphadistribution}
Conjugate pairs of $\alpha$-stable kernels $k$ and $\alpha$-stable distributions on $\mathbb{R}^d$.
\begin{enumerate}
\item For $\alpha=2$, let $k_{R}(x,y) = \frac{1}{\sqrt{(2 \pi)^d |R|}}\exp(-\frac{1}{2}(x-y)^{\top} R^{-1} (x-y))$ be a Gaussian kernel and $\mathcal{H}_{R}$ be its RKHS. Let $P$, $Q$ be two Gaussians $N(\mu_P, R_P)$ and $N(\mu_Q, R_Q)$, respectively. Then, the kernel mean is given by the Gaussian pdf $m_P =f_{\alpha}(\cdot |\mu_P, R+R_P)$ and the RKHS inner product is given by the Gaussian pdf ${\left\langle {{m_P},{m_Q}} \right\rangle _{\mathcal{H}_{R}}}= f(\mu_P|\mu_Q, R+R_P+R_Q)$.

\item For each $\alpha \in (0,2)$, let $k_{\alpha,\sigma}(x, y)=\kappa_{\alpha,\sigma}(x-y)$, $x,y \in \mathbb{R}$, be an $\alpha$-stable kernel on $\mathbb{R}$ and $\mathcal{H}_{\alpha,\sigma}$ be its RKHS. Let $P, Q$ be two $\alpha$-stable laws $S_{\alpha}(\sigma_P, \beta_P, \mu_P)$ and $S_{\alpha}(\sigma_Q, \beta_Q, \mu_Q)$, respectively, on $\mathbb{R}$. Then, the kernel mean is given by the stable pdf $m_P = f_{\alpha}(\cdot |(\sigma_P^\alpha+\sigma^\alpha)^{1/\alpha}, \frac{\beta_P \sigma_P^\alpha}{\sigma_P^\alpha+\sigma^\alpha}, \mu_P)$ and the RKHS inner product is given by the stable pdf ${\left\langle {{m_P},{m_Q}} \right\rangle _{\mathcal{H}_{\alpha, \sigma}}}= f_{\alpha}(\mu_P |(\sigma_P^\alpha+\sigma_Q^\alpha+ \sigma^\alpha)^{1/\alpha}, \frac{\beta_Q \sigma_Q^\alpha-\beta_P \sigma_P^\alpha}{\sigma_Q^\alpha+\sigma_P^\alpha +\sigma^\alpha}, \mu_Q)$. If $\alpha=1$ and $\beta=0$, then $S_{1}(\sigma, 0, \mu)$ corresponds to the Cauchy distribution.

\item For each $\alpha \in (0,2)$, let $k_{\alpha,\Gamma_s}(x, y)=\kappa_{\alpha,\Gamma_s}(x-y)$, $x,y \in \mathbb{R}^d$, be an $\alpha$-stable kernel on $\mathbb{R}^d$, where $\Gamma_s$ is a symmetric spectral measure, and let $\mathcal{H}_{\alpha,\Gamma_s}$ be its RKHS. Let $P, Q$ be two $\alpha$-stable laws $S_{\alpha}(\mu_P, \Gamma_P)$ and $S_{\alpha}(\mu_Q, \Gamma_Q)$, respectively, on $\mathbb{R}^d$. Then, the kernel mean is given by the stable pdf $m_P = f_{\alpha}(\cdot | \mu_P, \Gamma_P +\Gamma_s)$ and the RKHS inner product is given by the stable pdf ${\left\langle {{m_P},{m_Q}} \right\rangle _{\mathcal{H}_{\alpha, \sigma}}}= f_{\alpha}(\mu_P |\mu_Q, \Gamma_Q+ \tilde \Gamma_P+\Gamma_s)$.

\item For each $\alpha \in (0,2)$, let $k_{\alpha,R}(x, y)=\kappa_{\alpha,R}(x-y)$, $x,y \in \mathbb{R}^d$ be a sub-Gaussian $\alpha$-stable kernel on $\mathbb{R}^d$ and let $\mathcal{H}_{\alpha,R}$ be its RKHS. Let $P, Q \in \mathbb{SG}_{\alpha}(\mathbb{R}^{d})[R]$ be two sub-Gaussian $\alpha$-stable laws $S_{\alpha}(\mu_P, c_PR)$ and $S_{\alpha}(\mu_Q, c_QR)$, respectively, on $\mathbb{R}^d$. Then, the kernel mean is given by the sub-Gaussian pdf $m_P = f_{\alpha}(\cdot|\mu_P, (c_P^{\frac{\alpha}{2}}+1)^{\frac{2}{\alpha}}R)$ and the RKHS inner product is given by the sub-Gaussian pdf ${\left\langle {{m_P},{m_Q}} \right\rangle _{\mathcal{H}_{\alpha, R}}}= f_{\alpha}(\mu_P|\mu_Q, (c_P^{\frac{\alpha}{2}}+c_Q^{\frac{\alpha}{2}}+1)^{\frac{2}{\alpha}}R)$. If $\alpha=1$, then $S_{1}(\mu, R)$ corresponds to multivariate Cauchy distributions with pdf $f(x) \propto (1+||x-\mu   ||^2_{R^{-1}})^{-\frac{d+1}{2}}$. 
\item Tempered stable distributions can also be considered as examples \citep[Table 3.2, p. 77]{VolatilityClustering2011}.
\end{enumerate}
\end{Example}

\subsection{Generalized Hyperbolic Distributions} \label{sec:GHdistributionsbody}
GH distributions on $\mathbb{R}^d$ are a rich model class that includes, e.g., NIGs, hyperbolic distributions, VG distributions, Laplace distributions, Cauchy distributions, and Student's $t$ distributions, as special cases and limiting cases \citep{InfinitedivisibilityofthehyperbolicandgeneralizedinverseGaussiandistributions, Prause1999, Hammerstein2010}. A list of parametric models is found in, e.g., \citet[Table 1.1 p.4]{Prause1999}.
The GH and related models are applied, e.g., to mathematical finance \citep{LevyProcessesinFinance_PricingFinancialDerivatives, FinancialModellingwithJumpProcesses, Thevariancegammamodelforsharemarketreturns, Thevariancegammaprocessandoptionpricing, ProcessesofnormalinverseGaussiantype, Apparentscaling, Thefinestructureofassetreturns_anempiricalinvestigation}. The Mat\'ern kernel, often used in machine learning, is a special case of the VG distributions. A GH distribution is obtained by a {\it normal mean-variance mixture} of a generalized inverse Gaussian (GIG) distribution, which is a special case of the normal mean-variance mixture of the generalized $\Gamma$-convolution \citep{AnextensionofthenotionofageneralizedGamma-convolution}. The pdfs of GIG, GH, NIG, and VG distributions are presented in Appendix \ref{sec:GHclass}.

We start by introducing a normal mean-variance mixture distribution.
Let $N_d(\mu, \Delta)$ be a Gaussian distribution with mean vector $\mu \in \mathbb{R}^d$ and covariance matrix $\Delta \in \mathbb{P}^{d}$.
A {\it normal mean-variance mixture} distribution $P$ on $\mathbb{R}^{d}$ is given by
\begin{eqnarray*}
P(dx) = \int_{{\mathbb{R}^ + }}^{} {N_d(\mu  + y \beta ,y\Delta )(dx)G(dy)}, \hspace{2mm} \beta \in \mathbb{R}^d, \label{eq:normalmeanvariancemixturedistribution}
\end{eqnarray*}
where $G$ is a mixing probability measure on $\mathbb{R}^{+}$ \cite[Definition 2.4, p. 78]{Hammerstein2010}.
$P = N_d(\mu  + y \beta ,y\Delta ) \circ G$ denotes a simple notation.
The closure properties of the convolution and the infinite divisibility of $G$ are preserved as follows:
\begin{Lemma}{\cite[Lemma 2.5, p. 68]{Hammerstein2010}} \label{Lemma:PropertyNormalMeanVarianceMixture}
Let $\mathbb{G}$ be a class of probability distributions on $(\mathbb{R}^{+},\mathcal{B}^{+})$ and $G,G_1,G_2 \in \mathbb{G}$.
\begin{enumerate}
\item If $G = {G_1}*{G_2} \in \mathbb{G}$, then
\begin{eqnarray*}
({N_d}({\mu _1} + y\beta ,y\Delta ) \circ {G_1})*({N_d}({\mu _2} + y\beta ,y\Delta ) \circ {G_2}) = {N_d}({\mu _1} + {\mu _2} + y\beta ,y\Delta ) \circ G.
\end{eqnarray*}
\item If $G$ is infinitely divisible, then so is ${N_d}({\mu } + y\beta ,y\Delta) \circ G$.
\end{enumerate}
\end{Lemma}
A GH distribution on $\mathbb{R}^{d}$ is given by a normal mean-variance mixture with the GIG distribution:
\begin{eqnarray*}
G{H_d}(\lambda ,\alpha ,\beta ,\delta ,\mu ,\Delta ): = {N_d}(\mu  + y\Delta \beta ,y\Delta ) \circ GIG(\lambda ,\delta ,\sqrt {{\alpha ^2} -  ||\beta ||_\Delta ^2 } ),
\end{eqnarray*}
 where the parameters imply $\lambda \in \mathbb{R}$, shape parameter $\alpha > 0$, skewness parameter $\beta$, scaling parameter $\delta$, location parameter $\mu$, and p.d. matrix $\Delta \in \mathbb{P}^{d}$ (see Appendices \ref{sec:GIGdistributons} and \ref{sec:GHdistribution} for more details). A univariate GH distribution on $\mathbb{R}$ is given by letting $d=1$ and $\Delta=1$.

 The GH distribution contains the following subclasses and limiting cases. Their pdfs are found in Appendices \ref{sec:NIGdistribution}, \ref{sec:VGdistribution}, and \citet{Hammerstein2010}).
 \begin{enumerate}
 \item If $\lambda = -\frac{1}{2}$, then $G{H_d}(-\frac{1}{2} ,\alpha ,\beta ,\delta ,\mu ,\Delta )$) corresponds to the NIG distribution:
 \begin{eqnarray*}
 NIG_d(\alpha ,\beta ,\delta ,\mu ,\Delta): = {N_d}(\mu  + y\Delta \beta ,y\Delta ) \circ GIG(-\frac{1}{2} ,\delta ,\sqrt {{\alpha ^2} -  ||\beta ||_\Delta ^2 } ).
 \end{eqnarray*}
 \item If $\lambda = \frac{d+1}{2}$, then $G{H_d}(\frac{d+1}{2} ,\alpha ,\beta ,\delta ,\mu ,\Delta )$ corresponds to the hyperbolic distribution $HYP_d(\alpha, \beta, \delta, \mu, \Delta)$.
 \item If $\lambda >0$ and $\delta \rightarrow 0$, then $G{H_d}(\lambda >0 ,\alpha ,\beta , 0 ,\mu ,\Delta )$ corresponds to the VG distribution
  \begin{eqnarray*}
 VG_d(\lambda, \alpha ,\beta,\mu ,\Delta): = {N_d}(\mu  + y\Delta \beta ,y\Delta ) \circ Gamma(\lambda,\frac{{\alpha ^2} -  ||\beta ||_\Delta ^2 }{2} ),
 \end{eqnarray*}
 where $Gamma(\lambda, \gamma)$ is the Gamma distribution with the pdf $f(x)=\frac{\gamma^{\lambda}}{\Gamma(\lambda)} x^{\lambda-1}e^{-\gamma x}$. 
 Furthermore, if $\lambda = \frac{d+1}{2}$ (i.e., the above hyperbolic case), then $VG_d(\frac{d+1}{2}, \alpha ,\beta,\mu ,\Delta)$ corresponds to the skewed Laplace distribution
   \begin{eqnarray*}
 LAP_d(\alpha, \beta, \mu, \Delta): = {N_d}(\mu  + y\Delta \beta ,y\Delta ) \circ Gamma(\frac{d+1}{2},\frac{{\alpha ^2} -  ||\beta ||_\Delta ^2 }{2} ),
 \end{eqnarray*}
 with the pdf $f(x) \propto e^{- \alpha ||x-\mu ||_{\Delta^{-1}} + \langle \beta, x- \mu \rangle}$. We have seen the case of $d=1$ in Example \ref{eq:LaplaceVG_infinitelydivisible}.
  \item If $\lambda <0$, $\alpha \rightarrow 0$, and $\beta \rightarrow \mathbf{0}$, then $G{H_d}(\lambda <0 ,0 ,\mathbf{0}, \delta ,\mu ,\Delta )$ corresponds to the scaled and shifted $t$ distribution with $f = - 2 \lambda$ degrees of freedom:
  \begin{eqnarray*}
 t_d(\lambda, \delta,\mu ,\Delta): = {N_d}(\mu,y\Delta ) \circ iGamma(\lambda,\frac{\delta^2}{2} ),
 \end{eqnarray*}
 where $iGamma(\lambda, \delta)$ is the inverse Gamma distribution with the pdf $f(x)=\frac{x^{\lambda-1}}{\delta^{\lambda} \Gamma(-\lambda)} e^{- \frac{\delta}{x}}$.
  Furthermore, if $\lambda = -\frac{1}{2}$ (i.e., the above NIG case), then $t_d(-\frac{1}{2}, \delta,\mu ,\Delta)$ corresponds to the multivariate Cauchy distribution
 \begin{eqnarray*}
 CAU(\delta,\mu ,\Delta): = {N_d}(\mu,y\Delta ) \circ iGamma(-\frac{1}{2},\frac{\delta^2}{2} ),
 \end{eqnarray*}
 with the pdf $f(x) \propto (1+\frac{||x-\mu   ||^2_{\Delta^{-1}}}{\delta^2})^{-\frac{d+1}{2}}$, which is also shown in Example \ref{ex:conjugate_alphadistribution}.
 \end{enumerate}
 These classes have the following convolution properties, by using Lemma \ref{Lemma:PropertyNormalMeanVarianceMixture} and Proposition \ref{eq:GIGconvolutionstability}, which are the multivariate extensions of the univariate case {\cite[eq. (1.9), p. 14]{Hammerstein2010}}.

\begin{Proposition} \label{Pro:convolutionRelationsubclassmultivariate}
For each $d \ge 1$, there are the following convolution properties in the $d$-dimensional GH distributions:
\begin{enumerate}
\item \hspace{-0mm}$NIG_{d}(\alpha ,\beta ,{\delta _1},{\mu _1},\Delta)*NIG_{d}(\alpha ,\beta ,{\delta _2},{\mu _2},\Delta) = NIG_{d}(\alpha ,\beta ,{\delta _1} + {\delta _2},{\mu _1} + {\mu _2},\Delta)$,
\item \hspace{-0mm}$VG_{d}({\lambda _1},\alpha ,\beta, {\mu _1},\Delta)*VG_{d}({\lambda _2},\alpha ,\beta ,{\mu _2},\Delta) = VG_d({\lambda _1} + {\lambda _2},\alpha ,\beta ,{\mu _1} + {\mu _2},\Delta)$,
\item \hspace{-0mm}$NIG_{d}(\alpha ,\beta ,{\delta _1},{\mu _1},\Delta)*GH_{d}(1/2,\alpha ,\beta ,{\delta _2},{\mu _2},\Delta) = GH_{d}(1/2,\alpha ,\beta ,{\delta _1} + {\delta _2},{\mu _1} + {\mu _2},\Delta)$,
\item \hspace{-0mm}$GH_{d}( - \lambda ,\alpha ,\beta ,\delta ,{\mu _1},\Delta)*GH_{d}(\lambda ,\alpha ,\beta ,0,{\mu _2},\Delta) = GH_{d}(\lambda ,\alpha ,\beta ,\delta ,{\mu _1} + {\mu _2},\Delta)$,
\end{enumerate}
where $\lambda,\lambda_1,\lambda_2 >0$.
\end{Proposition}
These convolution properties can also be obtained by looking up their characteristic functions and L\'evy measures in \citet[Section 1.6.4, p. 46, Section 2.3, p. 79]{Hammerstein2010}.
Properties $1$ and $2$ imply a convolution semigroup. Property $3$ implies an absorbing property. Property $4$ implies another convolution property. By observing proposition \ref{Pro:convolutionRelationsubclassmultivariate}, we obtain the following conjugate, absorbing, and related pairs in GH kernels and GH distributions. 
The parametric models in Proposition \ref{Pro:convolutionRelationsubclassmultivariate} contain p.d. kernels $\kappa$ if and only if $\beta = \mathbf{0}$. Each example ($1-4$)  in the following corresponds to each property ($1-4$) in Proposition \ref{Pro:convolutionRelationsubclassmultivariate}. 

\begin{Example} \label{ex:conjugate_GHdistribution}
Conjugate, absorbing, and related pairs in the GH class.
\begin{enumerate}
\item Let $k_{\alpha, \delta, \Delta}(x,y)$ be a shift invariant NIG p.d. kernel and $\mathcal{H}_{\alpha, \delta, \Delta}$ be the RKHS. Let $P$, $Q$ be two NIG distributions $NIG(\alpha, \mathbf{0}, \delta_P, \mu_P, \Delta)$ and $NIG(\alpha, \mathbf{0}, \delta_Q, \mu_Q, \Delta)$, respectively. Then, the kernel mean is the NIG pdf $m_P =f(\cdot |\alpha, \mathbf{0}, \delta_P + \delta, \mu_P, \Delta)$ and the RKHS inner product is the NIG pdf ${\left\langle {{m_P},{m_Q}} \right\rangle _{\mathcal{H}_{\alpha, \delta, \Delta}}}= f(\mu_P|\alpha, \mathbf{0}, \delta_P+\delta_Q +\delta, \mu_Q, \Delta)$.
If $\alpha \rightarrow 0$, then these correspond to the Cauchy case.

\item Let $k_{\lambda, \alpha, \Delta}(x,y)$ be a shift invariant VG p.d. kernel\footnote{The Mat\'ern kernel corresponds to $\Delta=I$, and $\alpha=\frac{\sqrt{2 \nu}}{\sigma}$ \cite[Section 4.2.1]{GaussianProcessesforMachineLearning} \cite[p. 1533]{Sriperumbudur_JMLR2010}} and $\mathcal{H}_{\lambda, \alpha, \Delta}$ be the RKHS. Let $P$, $Q$ be two VG distributions $VG(\lambda_P, \alpha, \mathbf{0}, \mu_P, \Delta)$ and $VG(\lambda_Q, \alpha, \mathbf{0}, \mu_Q, \Delta)$, respectively. Then, the kernel mean is the VG pdf $m_P =f(\cdot |\lambda_P + \lambda, \alpha, \mathbf{0}, \mu_P, \Delta)$ and the RKHS inner product is the VG pdf ${\left\langle {{m_P},{m_Q}} \right\rangle _{\mathcal{H}_{\lambda, \alpha, \Delta}}}= f(\mu_P|\lambda_P + \lambda_Q+ \lambda, \alpha, \mathbf{0}, \mu_Q, \Delta)$. If $\lambda=\frac{d+1}{2}$, $\lambda_P=\frac{d+1}{2}$, or $\lambda_Q=\frac{d+1}{2}$, then these correspond to the Laplace case.

\item Let $k_{\alpha, \delta, \Delta}(x,y)$ be a NIG kernel and $\mathcal{H}_{\alpha, \delta, \Delta}$ be the RKHS. Let $P$ be a GH distribution $GH(1/2, \alpha, \mathbf{0}, \delta_P, \mu_P, \Delta)$. Then, the kernel mean is the GH pdf $m_P =f(\cdot |1/2, \alpha, \mathbf{0}, \delta_P + \delta, \mu_P, \Delta)$. If $\alpha \rightarrow 0$, then the NIG kernel $k_{0, \delta, \Delta}(x,y)$ corresponds to the Cauchy kernel.

Let $k_{1/2, \alpha, \delta, \Delta}(x,y)$ be a GH kernel and $\mathcal{H}_{1/2, \alpha, \delta, \Delta}$ be the RKHS. Let $P$, $Q$ be two NIG distributions $NIG(\alpha, 0, \delta_P, \mu_P, \Delta)$ and $NIG(\alpha, \mathbf{0}, \delta_Q, \mu_Q, \Delta)$, respectively. Then, the kernel mean is the GH pdf $m_P =f(\cdot |1/2, \alpha, \mathbf{0}, \delta_P + \delta, \mu_P, \Delta)$ and the RKHS inner product is the GH pdf ${\left\langle {{m_P},{m_Q}} \right\rangle _{\mathcal{H}_{1/2, \alpha, \delta, \Delta}}}= f(\mu_P|1/2, \alpha, \mathbf{0}, \delta_P+\delta_Q +\delta, \mu_Q, \Delta)$.
If $\alpha \rightarrow 0$, then the NIG distributions, $P$ and $Q$, correspond to the Cauchy distributions.

\item For $\lambda >0$, let $k_{-\lambda, \alpha, \delta, \Delta}(x,y)$ be a GH kernel and $\mathcal{H}_{-\lambda, \alpha, \delta, \Delta}$ be the RKHS. Let $P$ be a GH distribution $GH(\lambda, \alpha, \mathbf{0}, 0, \mu_P, \Delta)$. Then, the kernel mean is the GH pdf $m_P =f(\cdot |\lambda, \alpha, \mathbf{0}, \delta, \mu_P, \Delta)$.
If $\alpha \rightarrow 0$, then $k_{-\lambda, 0, \delta, \Delta}(x,y)$ corresponds to the Student's $t$ kernel. Furthermore, if $\lambda=\frac{1}{2}$, then $k_{-\frac{1}{2}, 0, \delta, \Delta}(x,y)$ corresponds to the Cauchy kernel.

For $\lambda >0$, let $k_{\lambda, \alpha, \Delta}(x,y)$ be a GH kernel and $\mathcal{H}_{\lambda, \alpha, \Delta}$ be the RKHS. Let $P$ be a GH distribution $GH(-\lambda, \alpha, \mathbf{0}, \delta_P, \mu_P, \Delta)$. Then, the kernel mean is the GH pdf $m_P =f(\cdot |\lambda, \alpha, \mathbf{0}, \delta_P, \mu_P, \Delta)$. If $\alpha \rightarrow 0$, then $P$ is the Student's $t$ distribution. Furthermore, if $\lambda=-\frac{1}{2}$, then $P$ is the Cauchy distribution.

\end{enumerate}
\end{Example}

\section{Connection to Machine Learning} \label{sec:ConnectionML}
As mentioned in the Introduction, absorbing and conjugate kernels (Examples \ref{ex:conjugate_alphadistribution} and \ref{ex:conjugate_GHdistribution}) provide a way to compute the RKHS values (i) $f(x)$, $x \in \mathbb{R}^d$, and the RKHS inner products (ii) ${\left\langle {{f},{g}} \right\rangle _{\mathcal{H}}} $ when $f, g \in \mathcal{H}$ are expressed by the weighted sums of parametric kernel means, $f=\sum_{i=1}^{n}w_i m_{P_i}$ and $g=\sum_{j=1}^{l} \tilde w_j m_{Q_j}$ for $\{P_i \}, \{Q_j \} \subset \mathcal{P}_{\Theta}$. Many algorithms aim to use the convolution trick. Examples include as follows:
\begin{itemize}

\item The difference between a probability measure $P \in \mathcal{M}_1(\mathbb{R}^d)$ and a model $P_{\theta} \in \mathcal{P}_{\Theta}$ in the RKHS norm $|| m_P- m_{P_{\theta}} ||_{\mathcal{H}}$ needs to be computed, e.g., for the purpose of a goodness-of-fit test and model criticism \citep{ModelCriticismNIPS2015}, based on the maximum mean discrepancy (MMD) \citep{DBLP:journals/jmlr/GrettonBRSS12}.

\item Various kernels $k(P,P_\theta)$ between a probabilistic measure $P$ and a model $P_\theta$, e.g., $k(P, P_{\theta})= \exp (-\frac{|| m_P- m_{P_{\theta}} ||^2_{\mathcal{H}}}{2 \sigma^2})$ need to be computed, as in the support measure machine \citep{NIPS2012_0015}.

\item \cite{TailoringICML2008} and \cite{McCalman2013} studied an approximation of a target probability measure $P \in \mathcal{M}_{1}(\mathbb{R}^d)$ with a Gaussian mixture model $P_{\theta} = \sum_{i=1}^{n} \theta_i P_i$ via solving the following optimization problem:
\begin{eqnarray*}
\hat \theta = \mathrm{argmin}_{\theta} ||m_{P}-m_{P_{\theta}}||^2_{\mathcal H} + \Omega(\theta) = \mathrm{argmin}_{\theta} ||m_{P}-\sum_{i=1}^n \theta_i m_{P_i}||^2_{\mathcal H} +\Omega(\theta),
\end{eqnarray*}
where $\Omega(\theta)$ is a regularization term, $\frac{\lambda}{2}|| \theta ||^{2}$ ($\lambda >0$). This optimization is solved by a constrained quadratic program: $\mathrm{min}_{\theta} \frac{1}{2} \theta^{\top} (A+ \lambda I_n) \theta - b^{\top} \theta$ subject to $\sum_{i=1}^n \theta_i =1$ and $ \theta \ge 0$, where we then need the computation of matrix $A \in \mathbb{R}^{n \times n}$ and vector $b \in \mathbb{R}^n$:
\begin{eqnarray*}
{A_{ij}} = {\left\langle {m_{P_i},m_{P_j}} \right\rangle _{\mathcal H}},  \hspace{2mm} {b_j} = {\left\langle {\hat m_P,m_{P_j}} \right\rangle _\mathcal{H}}, \hspace{3mm}  1 \le i,j \le n,
\label{eq:Qmatrixquadraticproblem}
\end{eqnarray*}
for parametric kernel means $\{ m_{P_i} \}$.

\item As mentioned in the Introduction, the kernel Bayesian inference  (KBI), which employs Bayesian inference in kernel mean form, has been proposed (\citealt{KernelBayes'Rule_BayesianInferencewithPositiveDefiniteKernels},  \citealt{KernelEmbeddingofConditionalDistributions}). KBI is applied to, e.g., filtering and smoothing algorithms on state space models (\citealt{KernelBayes'Rule_BayesianInferencewithPositiveDefiniteKernels} \citealt{KernelMonteCarlo2016}, \citealt{KernelBayesSmoothingAISTATS2016}) and policy learning in reinforcement learning (\citealt{DBLP:journals/corr/abs-1206-4655}, \citealt{DBLP:conf/uai/NishiyamaBGF12}, \citealt{PathIntegralControlbyReproducingKernelHilbertSpaceEmbedding2013},  \citealt{HilbertSpaceEmbeddingsofPredictiveStateRepresentations}).
When we extend it to {\it semiparametric} KBI, which combines nonparametric inference and parametric inference, we may want to use the RKHS functions $f=\sum_{i=1}^{n}w_i m_{P_{\theta_i}} \in \mathcal{H}$ expressed by parametric kernel means $\{ P_{\theta_i}\} \in \mathcal{P}_{\Theta}$, as is used in the model-based kernel sum rule (Mb-KSR) \citep{Nishiyama2014}.
\item Preimage algorithms \citep{Mika99kernelpca, KernelBayes'Rule_BayesianInferencewithPositiveDefiniteKernels} and kernel herding algorithms \citep{KernelHerdingUAI2010} can also be extended to estimators $f=\sum_{i=1}^{n}w_i m_{P_{\theta_i}}$ with parametric kernel means $\{P_{\theta_i}\}$. 
\end{itemize}

\section{Computation of Conjugate Kernels (Convolution Trick)} \label{sec:ComputationConvolutionTrick}
In Section \ref{sec:kernelmeansandIDD}, we mathematically investigated that several convolution tricks hold within a general convolution trick (Proposition \ref{Prop:CIDkernelmean}): the computation of kernel mean values and RKHS inner products is the same as the computation of p.d. kernels having different parameters, if conjugate kernels are used. However, conjugate kernels do not provide a tractable computation in general. We then discuss the computation of the conjugate kernels: $\alpha$-stable kernels and GH kernels.   

\begin{itemize}
\item It is known that $\alpha$-stable pdfs do not generally have a closed-form expression except for some special cases, Gaussians ($\alpha=2$) and Cauchy ($\alpha=1$), as given in Appendix \ref{sec:closedformsolutionunivariatestable}. Gaussian and Cauchy kernels may be used as tractable conjugate kernels. For other $\alpha$-stable kernels ($\alpha \neq 2$ and $\alpha \neq 1$), some numerical elaborations or approximations may be needed for the computation of the pdfs. The STABLE 5.1\footnote{John Nolan's Page. \url{http://academic2.american.edu/~jpnolan/stable/stable.html}} software allows the computation of $\alpha$-stable pdfs when they are independent, isotropic, elliptical, or have discrete spectral measures $\Gamma_d$ under some settings. More information can be found in the STABLE 5.1 software manual. For elliptically contoured $\alpha$-stable sub-Gaussian kernels on any dimension $\mathbb{R}^{d}$, the computation of pdfs is sufficient only to compute a one-dimensional amplitude function $\tilde \kappa(r)$ in equation (\ref{eq:radial_integral}), which can be computed by, e.g., a one-dimensional numerical integration. The STABLE 5.1 software supports the computation of sub-Gaussian pdfs in dimension $d < 100$. 

\item GH kernels and their subclasses are also elliptical pdfs, and the computation of the kernels is sufficient only to compute a one-dimensional amplitude function $\tilde \kappa(r)$. VG kernels or Mat\'ern kernels, which are a generalization of Laplace kernels, are used for covariance kernels in Gaussian processes. GH and NIG kernels are variants of Mat\'ern kernels, all of which are expressed by the Bessel function of the third kind. For example, there is an R package software called 'ghyp' on the GH distributions \citep{ghypGHdistribution2013}. 
\end{itemize}

In addition, random Fourier features \citep{RandomFeaturesNIPS2007} may be an approach to approximately compute conjugate kernels. From Proposition \ref{Prop:CIDkernelmean}, we have an equality
\begin{eqnarray*}
{\left\langle {{m_P},{m_Q}} \right\rangle _{\mathcal{H}_{A, \nu_s}}} = k_{A+A_P+A_Q,\nu_{s}+{\nu}_{P}+ \tilde{\nu}_{Q}}(\gamma_{P}, \gamma_{Q}) = \mathbb{E}_\omega[ \zeta_\omega(\gamma_{P}) \zeta_\omega(\gamma_{Q})^{*}].
\end{eqnarray*}
An RKHS inner product (l.h.s.) may be computed by approximating the expectation of $\zeta_\omega(\gamma_{P}) \zeta_\omega(\gamma_{Q})^{*}$ (r.h.s.) with sampling $\omega$ from the characteristic function having the generating triplet $(A+A_P+A_Q,\nu_{s}+{\nu}_{P}+ \tilde{\nu}_{Q})$.

\section{Conclusion} \label{sec:conclusion}
In this paper, we introduced a class of CID kernels that constitutes a large subclass in the set of shift-invariant characteristic kernels on $\mathbb{R}^d$, where CID kernels are closed under convolution but not closed under addition and pointwise product. We introduced absorbing, conjugate kernels, and convolution trick with respect to parametric models, where the basic computation of kernel mean values and RKHS inner products results in the computation of the same p.d. kernels with different parameters, which is an extension of kernel trick. Although the convolution trick may offer a mathematical view, the computation of conjugate kernels is not tractable in general. We then restrict convolution trick only to tractable cases or approximately compute intractable conjugate kernels. Future works include investigating the effectiveness of convolution trick in practice and developing approximation algorithms to efficiently compute intractable conjugate kernels.


\acks{We thank anonymous reviewers and the action editor for helpful comments. Y.N. thanks Prof. Tatsuhiko Saigo and Prof. Takaaki Shimura for a helpful discussion on infinitely divisible distributions. This work was supported in part by JSPS KAKENHI (grant nos. 26870821 and 22300098), the MEXT Grant-in-Aid for Scientific Research on Innovative Areas (no. 25120012), and by the Program to Disseminate Tenure Tracking System, MEXT, Japan.
}


\appendix

\section{$\alpha$-Stable Distributions} \label{sec:alpha-stableDist}
We briefly review the $\alpha$-stable distributions on $\mathbb{R}^d$.

\subsection{$\alpha$-Stable Distributions on $\mathbb{R}^{d}$} \label{sec:multivariatestabledistributionkernel}
The $\alpha$-stable distribution on $\mathbb{R}^{d}$ has the following characteristic function:
\begin{Theorem}{\cite[Theorem 2.3.1, p. 65]{Samorodnitsky1994}}
Let $\alpha \in (0,2)$. Then, $X=(X_1,\ldots,X_d)$ is an $\alpha$-stable random vector in $\mathbb{R}^{d}$ if and only if there exists a finite measure $\Gamma$ on the unit sphere $S_{d-1}=\{s \in \mathbb{R}^{d}: ||s||=1\}$ and a vector $\mu^{0} \in \mathbb{R}^{d}$ such that
\[ \hat P (\theta ) = \left\{ {\begin{array}{*{20}{c}}
{\exp \left( { - \int_{{S_{d - 1}}}^{} {|\theta^{\top}s{|^\alpha }\left( {1 - i{\mathop{\rm sgn}} (\theta^{\top}s)\tan \frac{{\pi \alpha }}{2}} \right)\Gamma (ds) + i} \theta^{\top} \mu ^0} \right)},&{(\alpha  \ne 1)}.\\
{\exp \left( { - \int_{{S_{d - 1}}}^{} {|\theta^{\top}s{|^\alpha }\left( {1 + i\frac{2}{\pi }{\mathop{\rm sgn}} (\theta^{\top}s)\ln |\theta^{\top}s|} \right)\Gamma (ds) + i} \theta^{\top} \mu ^0} \right)},&{(\alpha  = 1)}.
\end{array}} \right.\] 
The pair $(\Gamma,\mu^{0})$ is unique.
\end{Theorem}
The measure $\Gamma$ is called the \textit{spectral measure}.
See \citet[Section 2.3]{Samorodnitsky1994} for some examples of spectral measures. The radial sub-Gaussian distribution has a uniform spectral measure. An $\alpha$-stable random vector $X=(X_1,\ldots,X_d)$ has independent components if and only if its spectral measure $\Gamma$ is discrete and concentrated on the intersection of the axes with the sphere $S_{d-1}$.
It is known that any nondegenerate stable distribution on $\mathbb{R}^{d}$ has the $C^{\infty}$ pdf \cite[Example 28.2, p. 190]{Sato1999}.
An $\alpha$-stable distribution on $\mathbb{R}^d$ is symmetric if and only if $\mu^{0}=0$ and $\Gamma$ is a symmetric measure on $S_{d-1}$ (i.e., it satisfies $\Gamma(A)=\Gamma(-A)$ for any $A \in \mathcal{B}(S_{d-1})$) \cite[p.73]{Samorodnitsky1994}.

For each $\alpha \in (0,2)$, $\alpha$-stable distributions on $\mathbb{R}^d$ have the generating triplet $(0,\nu,\gamma)$ with
\begin{eqnarray}
\nu(B)= \int_{S_{d-1}} \Gamma(ds) \int_{0}^{\infty} 1_{B}(rs)\frac{dr}{r^{1+\alpha}},  \hspace{2mm} B \in \mathcal{B}(\mathbb{R}^d), \label{eq:stableLevymeasure}
\end{eqnarray}
where $\Gamma$ is the spectral measure on $S_{d-1}$ \cite[Theorem 14.3, p. 77]{Sato1999}. The sum of L\'evy measures $\nu_1 + \nu_2$ implies the sum of spectral measures $\Gamma_1 + \Gamma_2$.

\subsection{$\alpha$-Stable Distributions on $\mathbb{R}$} \label{sec:univariatestabledistributionkernel}
As a special case, an $\alpha$-stable distribution on $\mathbb{R}$ has the following characteristic function:
\begin{Theorem}{\cite[Definition 1.1.6, p. 5]{Samorodnitsky1994}}
A random variable $X$ is $\alpha$-stable ($\alpha \in (0,2]$) in $\mathbb{R}$ if and only if the parameters satisfy the conditions $\sigma \ge 0$, $\beta \in [-1,1]$, and $\mu \in \mathbb{R}$ such that its characteristic function has the form
\[\hat P(\theta) = \left\{ {\begin{array}{*{20}{c}}
{\exp \left( { - {\sigma ^\alpha }|\theta {|^\alpha }(1 - i\beta ({\mathop{\rm sgn}} \theta )\tan \frac{{\pi \alpha }}{2}) + i\mu \theta } \right)}&{(\alpha  \ne 1)},\\
{\exp \left( { - \sigma |\theta |(1 + i\beta \frac{2}{\pi }({\mathop{\rm sgn}} \theta )\ln |\theta |) + i\mu \theta } \right)}&{(\alpha  = 1)},
\end{array}} \right.\] where ${\mathop{\rm sgn}} \theta $ is a sign function
\[{\mathop{\rm sgn}} \theta  = \left\{ {\begin{array}{*{20}{c}}
1&{\theta  > 0,}\\
0&{\theta  = 0,}\\
{ - 1}&{\theta  < 0.}
\end{array}} \right.\]
When $\alpha \in (0,2)$, the parameters $\sigma$, $\beta$, and $\mu$ are unique. When $\alpha=2$, $\beta$ is irrelevant, and $\sigma$ and $\mu$ are unique.
\end{Theorem}
An $\alpha$-stable distribution on $\mathbb{R}$ is specified by the parameters $(\sigma, \beta, \mu)$, where $\sigma$ is a scale parameter, $\beta$ is a skewness parameter, and $\mu$ is a location parameter. $\sigma=0$ implies a delta measure. For $\alpha \in (0,2)$, an $\alpha$-stable distribution is symmetric if and only if $\beta=\mu=0$ \cite[Property 1.2.5, p. 11]{Samorodnitsky1994}.
A $2$-stable distribution is symmetric if and only if $\mu=0$.
An $\alpha$-stable density does not generally have a closed-form expression, except for some special cases. However, it is known that every nondegenerate stable distribution has the $C^{\infty}$ pdf \cite[Example 28.2, p. 190]{Sato1999}. Some known univariate $\alpha$-stable pdfs, expressed by elementary functions and special functions, are given in Appendix \ref{sec:closedformsolutionunivariatestable}.

The L\'evy measure $\nu$ of a univariate stable distribution is obtained by letting $d=1$ in the L\'evy measure (\ref{eq:stableLevymeasure}).
If $d=1$, then $S_0 =\{-1, 1\}$ and $\Gamma = \Gamma(\{-1\})\delta_{-1} + \Gamma(\{1 \}) \delta_{1}$, where $\Gamma(\{-1\}), \Gamma(\{1\}) \ge 0$ and $\Gamma(\{-1\}) + \Gamma(\{1\}) > 0$ {\cite[Example 2.3.3, p. 67]{Samorodnitsky1994}}. By substituting this into equation (\ref{eq:stableLevymeasure}), we can obtain the L\'evy measure $\nu$ of a univariate stable distribution as
\begin{eqnarray*}
\nu(dx) = \Gamma(\{1\}) \frac{1 }{x^{1+ \alpha}} 1_{(0,\infty)}(x)dx +\Gamma(\{-1\}) \frac{1 }{|x|^{1+ \alpha}} 1_{(-\infty,0)}(x)dx.
\end{eqnarray*}
A stable distribution $S_{\alpha}(\sigma, \beta, \mu)$ is given with the spectral measure as
\begin{eqnarray*}
\sigma = (\Gamma(\{1 \})+ \Gamma(\{-1 \}))^{\frac{1}{\alpha}} >0,  \hspace{2mm} \beta = \frac{(\Gamma(\{1 \})- \Gamma(\{-1 \}))}{\Gamma(\{1 \})+ \Gamma(\{-1 \})}  \in [-1,1].
\end{eqnarray*}

The sum of L\'evy measures $\nu_1 + \nu_2$ implies the sum of mass functions $\Gamma_1(\{-1 \}) + \Gamma_2(\{-1 \})$ and $\Gamma_1(\{1 \}) + \Gamma_2(\{1 \})$.
We can see the convolution property $S_{\alpha}(\sigma_1, \beta_1, \mu_1) * S_{\alpha}(\sigma_2, \beta_2, \mu_2) = S_{\alpha}((\sigma_1^{\alpha} +\sigma_2^{\alpha})^{\frac{1}{\alpha}}, \frac{\sigma_1^{\alpha}\beta_1 +\sigma_2^{\alpha}\beta_2 }{\sigma_1^{\alpha} +\sigma_2^{\alpha}}, \mu_1+\mu_2)$ of the univariate stable distribution from the viewpoint of the spectral measure.

\subsection{Closed-Form and Special Function Form of $\alpha$-Stable PDFs on $\mathbb{R}$} \label{sec:closedformsolutionunivariatestable}
There are three cases where the $\alpha$-stable pdf on $\mathbb{R}$ is expressed by elementary functions:
\begin{enumerate}
\item The $2$-stable distribution $S_{2}(\sigma, \beta, \mu)$ is the Gaussian $N(\mu, 2\sigma^2)$, where $\beta$ has no effect, with the pdf
\begin{eqnarray*}
{f_{Gauss}}(x) = \frac{1}{{2\sigma \sqrt \pi  }}{e^{ - \frac{{{{(x - \mu )}^2}}}{{4{\sigma ^2}}}}},x \in \mathbb{R}. \label{eq:Gaussexample}
\end{eqnarray*}
\item The $1$-stable distribution $S_{1}(\sigma, \beta=0, \mu)$ is the Cauchy distribution with the pdf
\begin{eqnarray*}
{f_{Cauchy}}(x) = \frac{\sigma }{{\pi ({{(x - \mu )}^2} + {\sigma ^2})}},x \in \mathbb{R}. \label{eq:Cauchyexample}
\end{eqnarray*}
\item The $1/2$-stable distribution $S_{1/2}(\sigma, \beta= \pm 1, \mu)$ is the L\'evy distribution with the pdf
\begin{eqnarray*}
{f_{Levy}}(x) = \frac{{\sqrt \sigma  }}{{\sqrt {2\pi } {{(x - \mu )}^{3/2}}}}{e^{ - \frac{\sigma }{{2(x - \mu )}}}},\mu  < x < \infty. \label{eq:Levyexample}
\end{eqnarray*}
\end{enumerate}

There are some cases where the $\alpha$-stable pdf is expressed by special functions.
The following expression is found in \citet{Continuousanddiscretepropertiesofstochasticprocesses}.
Note that kernel means $m_P$ and RKHS inner products also take these expressions. For simplicity, we only show standardized stable pdfs ${d_{stable}}(x;\alpha ,\sigma=1,\beta ,\mu=0)$.

\vspace{2mm}

\noindent {\bf Fresnel integrals:}

If $(\alpha,\sigma ,\beta,\mu)=(1/2,1,0,0)$,
\begin{eqnarray*}
\hspace{-10mm}&&{d_{stable}}(x;1/2 ,1,0,0) \\
\hspace{-10mm}&&= \frac{{|x{|^{ - \frac{3}{2}}}}}{{\sqrt {2\pi } }}\left( {\sin \left( {\frac{1}{{4|x|}}} \right)\left( {\frac{1}{2} - S\left( {\sqrt {\frac{1}{{2\pi |x|}}} } \right)} \right) + \cos \left( {\frac{1}{{4|x|}}} \right)\left( {\frac{1}{2} - C\left( {\sqrt {\frac{1}{{2\pi |x|}}} } \right)} \right)} \right),
\end{eqnarray*}
where $C(z)$ and $S(z)$ are the Fresnel integrals
\begin{eqnarray*}
C(z) = \int_0^z {\cos \left( {\frac{{\pi {t^2}}}{2}} \right)} dt, \hspace{5mm}  S(z) = \int_0^z {\sin \left( {\frac{{\pi {t^2}}}{2}} \right)} dt.
\end{eqnarray*}
This is a symmetric stable pdf. $k(x,y)={d_{stable}}(x-y;1/2 ,1,0,0)$, $x, y \in \mathbb{R}$, gives a characteristic p.d. kernel.

\vspace{2mm}

\noindent {\bf Modified Bessel function:}

If $(\alpha,\sigma ,\beta,\mu)=(1/3,1,1,0)$, the one-sided continuous density is
\begin{eqnarray*}
{d_{stable}}(x;1/3,1,1,0) = \frac{1}{\pi }\frac{{{2^{3/2}}}}{{{3^{7/4}}}}{x^{ - 3/2}}{K_{1/3}}\left( {\frac{{{2^{5/2}}}}{{{3^{9/4}}}}{x^{ - 1/2}}} \right),x \ge 0,
\end{eqnarray*}
where $K_{\nu}(x)$ is a modified Bessel function of the third kind.

\vspace{2mm}

\noindent {\bf Hypergeometric function:}

If $(\alpha,\sigma ,\beta,\mu)=(4/3,1,0,0)$,
\begin{eqnarray*}
{d_{stable}}(x;\frac{4}{3},1,0,0) &=& \frac{{{3^{5/4}}\Gamma (7/12)\Gamma (11/12)}}{{{2^{5/2}}\sqrt \pi  \Gamma (6/12)\Gamma (8/12)}}{_2F_2}\left( {\frac{7}{{12}},\frac{{11}}{{12}};\frac{6}{{12}},\frac{8}{{12}};\frac{{{3^3}{x^4}}}{{{2^8}}}} \right) \\ \nonumber
&&- \frac{{{3^{11/4}}{{\left| x \right|}^3}\Gamma (13/12)\Gamma (17/12)}}{{{2^{13/2}}\sqrt \pi  \Gamma (18/12)\Gamma (15/12)}}{_2F_2}\left( {\frac{{13}}{{12}},\frac{{17}}{{12}};\frac{{18}}{{12}},\frac{{15}}{{12}};\frac{{{3^3}{x^4}}}{{{2^8}}}} \right), x \in \mathbb{R},
\end{eqnarray*}
where $_p{F_q}$ is the (generalized) hypergeometric function
\begin{eqnarray*}
{}_p{F_q}({a_1}, \ldots ,{a_p};{b_1}, \ldots ,{b_q};z) = \sum\limits_{n = 0}^\infty  {\frac{{{{({a_1})}_n} \cdots {{({a_p})}_n}}}{{{{({b_1})}_n} \cdots {{({b_q})}_n}}}} \frac{{{z^n}}}{{n!}}
\end{eqnarray*}
with the Pochhammer symbol $(a)_{0}=1$, $(a)_n=a(a+1)\ldots(a+n-1)$ for $n \in \mathbb{N}^{+}$.
This is a symmetric stable pdf. $k(x,y)={d_{stable}}(x-y;\frac{4}{3},1,0,0)$, $x, y \in \mathbb{R}$, gives a characteristic p.d. kernel.

If $(\alpha,\sigma ,\beta,\mu)=(3/2,1,0,0)$ (the Holtsmark distribution),
\begin{eqnarray*}
{d_{stable}}(x;\frac{3}{2},1,0,0) &=& \frac{1}{\pi }\Gamma (5/3){}_2{F_3}\left( {\frac{5}{{12}},\frac{{11}}{{12}};\frac{1}{3},\frac{1}{2},\frac{5}{6}; - \frac{{{2^2}{x^6}}}{{{3^6}}}} \right)\\ \nonumber
&& - \frac{{{x^2}}}{{3\pi }}{}_3{F_4}\left( {\frac{3}{4},1,\frac{5}{4};\frac{2}{3},\frac{5}{6},\frac{7}{6},\frac{4}{3}; - \frac{{{2^2}{x^6}}}{{{3^6}}}} \right)\\ \nonumber
&& + \frac{{7{x^4}}}{{{3^4}\pi }}\Gamma (4/3){}_2{F_3}\left( {\frac{{13}}{{12}},\frac{{19}}{{12}};\frac{7}{6},\frac{3}{2},\frac{5}{3}; - \frac{{{2^2}{x^6}}}{{{3^6}}}} \right), x \in \mathbb{R}.
\end{eqnarray*}
This is a symmetric stable pdf. The Holtsmark kernel $k(x,y)={d_{stable}}(x-y;3/2,1,0,0)$, $x,y \in \mathbb{R}$, gives a characteristic p.d. kernel.

\vspace{2mm}

\noindent {\bf Whittaker function:}

If $(\alpha,\sigma ,\beta,\mu)=(2/3,1,0,0)$,
\begin{eqnarray*}
{d_{stable}}(x;2/3,1,0,0) = \frac{1}{{2\sqrt {3\pi } \left| x \right|}}\exp \left( {\frac{2}{{27{x^2}}}} \right){W_{ - 1/2,1/6}}\left( {\frac{4}{{27{x^2}}}} \right),x \in \mathbb{R},
\end{eqnarray*}
where ${W_{\lambda ,\mu }}(z)$ is the Whittaker function defined as
\begin{eqnarray*}
{W_{\lambda ,\mu }}(z) &=& \frac{{{z^\lambda }{e^{ - z/2}}}}{{\Gamma (\mu  - \lambda  + 1/2)}}\int_0^\infty  {{{\mathop{\rm e}\nolimits} ^{ - t}}{t^{\mu  - \lambda  - 1/2}}{{\left( {1 + \frac{t}{z}} \right)}^{\mu  - \lambda  - 1/2}}dt,} \\ \nonumber
&& {\mathop{\rm Re}\nolimits} (\mu  - \lambda ) >  - \frac{1}{2},\left| {\arg (z)} \right| < \pi.
\end{eqnarray*}
This is a symmetric stable pdf. $k(x,y)={d_{stable}}(x-y;2/3,1,0,0)$, $x,y \in \mathbb{R}$, gives a characteristic p.d. kernel.

If $(\alpha,\sigma ,\beta,\mu)=(2/3,1,1,0)$, the one-sided density is
\begin{eqnarray*}
{d_{stable}}(x;2/3,1,1,0) = \sqrt {\frac{3}{\pi }} \frac{1}{{\left| x \right|}}\exp \left( { - \frac{{16}}{{27{x^2}}}} \right){W_{1/2,1/6}}\left( {\frac{{32}}{{27{x^2}}}} \right),x \ge 0.
\end{eqnarray*}

If $(\alpha,\sigma ,\beta,\mu)=(3/2,1,1,0)$, the $\alpha$-stable density is
\begin{eqnarray*}
{d_{stable}}(x;2/3,1,1,0) = \left\{ {\begin{array}{*{20}{c}}
{\sqrt {\frac{3}{\pi }} \frac{1}{{\left| x \right|}}\exp \left( {\frac{{{x^3}}}{{27}}} \right){W_{1/2,1/6}}\left( { - \frac{2}{{27}}{x^3}} \right),}&{x < 0}\\
{\frac{1}{{2\sqrt {3\pi } \left| x \right|}}\exp \left( {\frac{{{x^3}}}{{27}}} \right){W_{ - 1/2,1/6}}\left( {\frac{2}{{27}}{x^3}} \right),}&{x > 0}
\end{array}} \right.
\end{eqnarray*}

\vspace{2mm}

\noindent {\bf Lommel function:}

If $(\alpha,\sigma ,\beta,\mu)=(1/3,1,0,0)$,
\begin{eqnarray*}
{d_{stable}}(x;1/3,1,0,0) = {\mathop{\rm Re}\nolimits} \left( {\frac{{2\exp ( - i\pi /4)}}{{3\sqrt 3 \pi |x{|^{3/2}}}}{S_{0,1/3}}\left( {\frac{{2\exp (i\pi /4)}}{{3\sqrt 3 |x{|^{1/2}}}}} \right)} \right).
\end{eqnarray*}
Here, the Lommel functions $s_{\mu, v}(z)$ and $S_{\mu,v}(z)$ are defined by
\begin{eqnarray*}
{s_{\mu ,v}}(z) &=& \frac{\pi }{2}\left( {{Y_\nu }(z)\int_0^z {{z^\mu }{J_v}(z)dz}  - {J_\nu }(z)\int_0^z {{z^\mu }{Y_v}(z)dz} } \right),\\
{S_{\mu ,v}}(z) &=& {s_{\mu ,v}}(z) - \frac{{{2^{\mu  - 1}}\Gamma \left( {(1 + \mu  + \nu )/2} \right)}}{{\pi \Gamma \left( {(\nu  - \mu )/2} \right)}}\left( {{J_\nu }(z) - \cos \left( {\frac{{\mu  - \nu }}{2}\pi } \right){Y_v}(z)} \right),
\end{eqnarray*}
where $J_{\nu}(z)$ and $Y_{\nu}(z)$ are Bessel functions of the first and second kind, respectively.
This is a symmetric stable pdf. $k(x,y)={d_{stable}}(x-y;1/3,1,0,0)$, $x,y \in \mathbb{R}$, gives a characteristic p.d. kernel.

\vspace{2mm}

\noindent {\bf Landau distribution:}

If $(\alpha,\sigma ,\beta,\mu)=(1,1,1,0)$ (the Landau distribution),
\begin{eqnarray*}
{d_{stable}}(x;1,1,1,0) = \frac{1}{\pi }\int_0^\infty  {{e^{ - t\log t - xt}}\sin (\pi t)dt}.
\end{eqnarray*}

\subsection{Sub-Gaussian (Elliptically Contoured) $\alpha$-Stable Distributions on $\mathbb{R}^d$} \label{sec:subGaussclass}
The sub-Gaussian $\alpha$-stable distribution has the following characteristic function:

\begin{Proposition}{\cite[Proposition 2.5.2, p. 78]{Samorodnitsky1994}} \label{Prop:SubGaussianCharacteristicfunction}
Let $\alpha \in (0,2)$. The sub-Gaussian $\alpha$-stable random vector $X$ in $\mathbb{R}^{d}$ has the characteristic function
\begin{eqnarray*}
E\exp \left[ {i\sum\limits_{k = 1}^d {{\theta _k}{X_k}} } \right] = \exp \left( { - {{\left| {\frac{1}{2}\sum\limits_{ij = 1}^d {{\theta _i}{\theta _j}{R_{ij}}} } \right|}^{\frac{\alpha }{2}}}}+i(\theta,\mu^{0}) \right), \label{eq:SubGaussCharacteristicFunc}
\end{eqnarray*}
where $R$ is a p.d. matrix and $\mu^{0} \in \mathbb{R}^{d}$ is a shift vector.
\end{Proposition}
$\alpha=2$ and $\alpha=1$ imply the multivariate Gaussian and Cauchy distribution, respectively.

For $\alpha \in (0,2)$, the radial sub-Gaussian $\mathbb{SG}_{\alpha}(\mathbb{R}^d)[I]$ (with identity matrix $R = I$) has the uniform spectral measure $\Gamma(B) = c|B|$, $\forall B \in \mathcal{B}(S_{d-1})$ in the L\'evy measure (\ref{eq:stableLevymeasure}) {\cite[Proposition 2.5.5, p. 79]{Samorodnitsky1994}}. Sub-Gaussian $\mathbb{SG}_{\alpha}(\mathbb{R}^d)[R]$ with a p.d. matrix $R$ is the elliptical version of the radial sub-Gaussians. Its spectral measure is given in {\citet[Proposition 2.5.8, p. 82]{Samorodnitsky1994}}. 

\section{GH Classes on $\mathbb{R}^d$} \label{sec:GHclass}
A GH distribution on $\mathbb{R}^{d}$ is given by the normal mean-variance mixture with the GIG mixing distribution. See, e.g., \citet{Hammerstein2010} for more information. We here reproduce some of them.
\subsection{GIG Distributions on $\mathbb{R}^+$} \label{sec:GIGdistributons}
A generalized inverse Gaussian (GIG) distribution $GIG(\lambda ,\delta ,\gamma)$ on $\mathbb{R}^{+}$ is given by the following pdf:
\begin{eqnarray*}
{d_{GIG(\lambda ,\delta ,\gamma )}}(x) = {\left( {\frac{\gamma }{\delta }} \right)^\lambda }\frac{1}{{2{K_\lambda }(\delta \gamma )}}{x^{\lambda  - 1}}\exp \left( { - \frac{1}{2}\left( {\frac{{{\delta ^2}}}{x} + {\gamma ^2}x} \right)} \right){1_{(0,\infty )}}(x),
\end{eqnarray*}
where $K_{\lambda}(x)$ is the modified Bessel function of the third kind with index $\lambda$.
The parameters $(\lambda ,\delta ,\gamma )$ take the following values:
\[\left\{ {\begin{array}{*{20}{c}}
{\delta  \ge 0,\gamma  > 0,}& {\rm if} \hspace{3mm} {\lambda  > 0},\\
{\delta  > 0,\gamma  > 0,}&{\rm if} \hspace{3mm} {\lambda  = 0},\\
{\delta  > 0,\gamma  \ge 0,}&{\rm if} \hspace{3mm} {\lambda  < 0},
\end{array}} \right.\]
where $\delta=0$ and $\gamma=0$ correspond to limiting cases,\footnote{
If $\lambda  \ne 0$, then ${{K_\lambda }(x) \sim \frac{1}{2}\Gamma (|\lambda |){{\left( {\frac{x}{2}} \right)}^{ - |\lambda |}}}$ ($x \downarrow 0$).} which are the Gamma distribution and the inverse Gamma distribution, respectively. The GIG distributions have the following convolution properties:
\begin{Proposition}{\cite[Proposition 1.11, p. 11]{Hammerstein2010}} \label{eq:GIGconvolutionstability}
Within the class of GIG distributions, the following convolution properties hold:
\begin{itemize}
\item[a)] $GIG( - \frac{1}{2},{\delta _1},\gamma )*GIG( - \frac{1}{2},{\delta _2},\gamma ) = GIG( - \frac{1}{2},{\delta _1} + {\delta _2},\gamma )$,
\item[b)] $GIG( - \frac{1}{2},{\delta _1},\gamma )*GIG(\frac{1}{2},{\delta _2},\gamma ) = GIG(\frac{1}{2},{\delta _1} + {\delta _2},\gamma )$,
\item[c)] $GIG( - \lambda ,\delta ,\gamma )*GIG(\lambda ,0,\gamma ) = GIG(\lambda ,\delta ,\gamma )$, \hspace{15mm} $\lambda>0$,
\item[d)] $GIG({\lambda _1},0,\gamma )*GIG({\lambda _2},0,\gamma ) = GIG({\lambda _1} + {\lambda _2},0,\gamma )$, \hspace{5mm} $\lambda_1, \lambda_2 >0$.
\end{itemize}
\end{Proposition}

\subsection{GH Distributions on $\mathbb{R}^d$} \label{sec:GHdistribution}
A GH distribution has the following pdf:
\begin{eqnarray*}
\hspace{-10mm} &&{d_{GH_d(\lambda ,\alpha ,\beta ,\delta ,\mu ,\Delta )}}(x) = \nonumber \\
\hspace{-10mm} && a(\lambda ,\alpha ,\beta ,\delta ,\mu ,\Delta ){\left( {\sqrt {{\delta ^2} + ||x - \mu ||_{{\Delta ^{ - 1}}}^2} } \right)^{\lambda  - \frac{d}{2}}}{K_{\lambda  - \frac{d}{2}}}\left( {\alpha \sqrt {{\delta ^2} + ||x - \mu ||_{{\Delta ^{ - 1}}}^2} } \right){e^{\left\langle {\beta ,x - \mu } \right\rangle }},
\end{eqnarray*}
where $a(\lambda ,\alpha ,\beta ,\delta ,\mu ,\Delta )$ is the normalization constant:
\begin{eqnarray*}
a(\lambda ,\alpha ,\beta ,\delta ,\mu ,\Delta ) = \frac{{{{({\alpha ^2} - ||\beta ||_\Delta ^2)}^{\lambda /2}}}}{{{{(2\pi )}^{d/2}{|\Delta|^{\frac{1}{2}}}}{\alpha ^{\lambda  - d/2}}{\delta ^\lambda }{K_\lambda }(\delta \sqrt {{\alpha ^2} - ||\beta ||_\Delta ^2} )}}.
\end{eqnarray*}
The GH parameters $(\lambda ,\alpha ,\beta ,\delta ,\mu ,\Delta )$ take the following values:
\begin{eqnarray*}
\lambda  \in \mathbb{R},\hspace{2mm} \alpha ,\delta  \in {\mathbb{R}_+ },\hspace{2mm} \beta ,\mu  \in {\mathbb{R}^d},\hspace{2mm} \Delta \in \mathbb{P}_{d}, \hspace{2mm} \begin{array}{*{20}{c}}
{\delta  \ge 0,0 \le ||\beta ||_\Delta  < \alpha ,}&{\mathrm{if} \hspace{2mm}\lambda  > 0,}\\
{\delta  > 0,0 \le  ||\beta ||_\Delta  < \alpha ,}&{\mathrm{if} \hspace{2mm}\lambda  = 0,}\\
{\delta  > 0,0 \le ||\beta ||_\Delta  \le \alpha ,}&{\mathrm{if} \hspace{2mm}\lambda  < 0,}
\end{array} \label{eq:parameterRange}
\end{eqnarray*}
where $\delta=0$ or ${{\alpha} = ||\beta ||_\Delta}$ is a limiting case.
The GH distribution is symmetric if and only if $\beta=\mathbf{0}$ and $\mu=0$.
The symmetric GH has the following elliptical pdf:
\begin{eqnarray*}
{d_{SG{H_d}(\lambda ,\alpha ,\delta ,\Delta )}}(x) = \frac{{{\alpha ^{\frac{d}{2}}}}}{{{{(2\pi )}^{\frac{d}{2}}}{|\Delta|^{\frac{1}{2}}}{\delta ^\lambda }{K_\lambda }(\delta \alpha )}}{\left( {\sqrt {{\delta ^2} + ||x||_{{\Delta ^{ - 1}}}^2} } \right)^{\lambda  - \frac{d}{2}}}{K_{\lambda  - \frac{d}{2}}}\left( {\alpha \sqrt {{\delta ^2} + ||x||_{{\Delta ^{ - 1}}}^2} } \right),
\end{eqnarray*}
where $\nu(t)$ in equation (\ref{eq:radial_integral}) is given by a GIG distribution.

\subsection{NIG Distributions on $\mathbb{R}^d$} \label{sec:NIGdistribution}
The NIG distribution $NIG_d(\alpha, \beta, \delta, \mu, \Delta)$ has the following pdf \cite[p.74]{Hammerstein2010}:
\begin{eqnarray*}
{d_{NIG_d(\alpha ,\beta ,\delta ,\mu ,\Delta )}}(x) \propto
 {\left( {\sqrt {{\delta ^2} + ||x - \mu ||_{{\Delta ^{ - 1}}}^2} } \right)^{- \frac{d+1}{2}}}{K_{\frac{d+1}{2}}}\left( {\alpha \sqrt {{\delta ^2} + ||x - \mu ||_{{\Delta ^{ - 1}}}^2} } \right){e^{\left\langle {\beta ,x - \mu } \right\rangle }}.
\end{eqnarray*}

\subsection{VG Distributions on $\mathbb{R}^d$} \label{sec:VGdistribution}
The VG distribution $VG_d(\lambda, \alpha, \beta, \mu, \Delta)$ has the following pdf \cite[p.74]{Hammerstein2010}:\footnote{The VG pdf is bounded at $x=\mu$ if and only if $ \lambda > \frac{d}{2}$.}
\begin{eqnarray*}
{d_{VG_d(\lambda, \alpha ,\beta,\mu ,\Delta )}}(x) \propto
 {\left( { { ||x - \mu ||_{{\Delta ^{ - 1}}}} } \right)^{\lambda -\frac{d}{2}}}{K_{\lambda -\frac{d}{2}}}\left( {\alpha ||x - \mu ||_{{\Delta ^{ - 1}}}} \right){e^{\left\langle {\beta ,x - \mu } \right\rangle }}.
\end{eqnarray*}

\bibliography{JMLR_CIDkernel_revise}

\begin{thebibliography}{59}
\providecommand{\natexlab}[1]{#1}
\providecommand{\url}[1]{\texttt{#1}}
\expandafter\ifx\csname urlstyle\endcsname\relax
  \providecommand{\doi}[1]{doi: #1}\else
  \providecommand{\doi}{doi: \begingroup \urlstyle{rm}\Url}\fi

\bibitem[Applebaum(2009)]{Applebaum2009}
D.~Applebaum.
\newblock \emph{L\' evy processes and stochastic calculus}.
\newblock second edition, Cambridge University Press, 2009.

\bibitem[Aronszajn(1950)]{Aronszajn1950}
N.~Aronszajn.
\newblock {Theory of Reproducing Kernels}.
\newblock \emph{Transactions of the American Mathematical Society},
  68(3):\penalty0 337--404, 1950.

\bibitem[Barndorff-Nielsen(1998)]{ProcessesofnormalinverseGaussiantype}
E.~O. Barndorff-Nielsen.
\newblock Processes of normal inverse gaussian type.
\newblock \emph{Finance and Stochastics}, 2:\penalty0 41--68, 1998.

\bibitem[Barndorff-Nielsen and Prause(2001)]{Apparentscaling}
E.~O. Barndorff-Nielsen and K.~Prause.
\newblock Apparent scaling.
\newblock \emph{Finance and Stochastics}, 5:\penalty0 103--113, 2001.

\bibitem[Barndorff-Nielsen and
  Halgreen(1977)]{InfinitedivisibilityofthehyperbolicandgeneralizedinverseGaussiandistributions}
O.~E. Barndorff-Nielsen and C.~Halgreen.
\newblock Infinite divisibility of the hyperbolic and generalized inverse
  gaussian distributions.
\newblock \emph{Zeitschrift f\"ur Wahrscheinlichkeitstheorie und verwandte
  Gebiete}, 38:\penalty0 309--312, 1977.

\bibitem[Barndorff-Nielsen and
  Halgreen(1990)]{Thevariancegammamodelforsharemarketreturns}
O.~E. Barndorff-Nielsen and C.~Halgreen.
\newblock The variance gamma (v.g.) model for share market returns.
\newblock \emph{Journal of Business}, 63:\penalty0 511--524, 1990.

\bibitem[Berg et~al.(1984)Berg, Christensen, and Ressel]{BergChristensen1984}
C.~Berg, J.~P.~R. Christensen, and P.~Ressel.
\newblock \emph{Harmonic Analysis on Semigroups}.
\newblock Springer, 1984.

\bibitem[Berlinet and Thomas-Agnan(2004)]{BerlinetBook2004}
A.~Berlinet and C.~Thomas-Agnan.
\newblock \emph{Reproducing kernel Hilbert spaces in probability and
  statistics}.
\newblock Kluwer Academic Publisher, 2004.

\bibitem[Bianchi et~al.(2010)Bianchi, Rachev, Kim, and Fabozzi]{Bianchi2010}
M.~L. Bianchi, S.T. Rachev, Y.S. Kim, and F.J. Fabozzi.
\newblock Tempered infinitely divisible distributions and processes.
\newblock \emph{Theory of Probability and Its Applications (TVP), Society for
  Industrial and Applied Mathematics (SIAM)}, 55\penalty0 (1):\penalty0 59--86,
  2010.

\bibitem[Bochner(1959)]{Bochner1959}
S.~Bochner.
\newblock Lectures on fourier integrals. with an author's supplement on
  monotonic functions, stieltjes integrals, and harmonic analysis.
\newblock In \emph{Princeton University Press, Princeton, NJ}. 1959.

\bibitem[Boots et~al.(2013)Boots, Gordon, and
  Gretton]{HilbertSpaceEmbeddingsofPredictiveStateRepresentations}
B.~Boots, G.~Gordon, and A.~Gretton.
\newblock Hilbert space embeddings of predictive state representations.
\newblock \emph{Uncertainty in Artificial Intelligence (UAI)}, 2013.

\bibitem[Breymann and L\"uthi(2013)]{ghypGHdistribution2013}
W.~Breymann and D.~L\"uthi.
\newblock {ghyp: A package on generalized hyperbolic distributions}.
\newblock 2013.

\bibitem[Carr et~al.(2002)Carr, Geman, Madan, and
  Yor]{Thefinestructureofassetreturns_anempiricalinvestigation}
P.~Carr, H.~Geman, D.~B. Madan, and M.~Yor.
\newblock The fine structure of asset returns: an empirical investigation.
\newblock \emph{Journal of Business}, 75:\penalty0 305--332, 2002.

\bibitem[Chen et~al.(2010)Chen, Welling, and Smola]{KernelHerdingUAI2010}
Y.~Chen, M.~Welling, and A.~Smola.
\newblock {Super-Samples from Kernel Herding}.
\newblock In \emph{Uncertainty in Artificial Intelligence (UAI)}. 2010.

\bibitem[Cont and Tankov(2004)]{FinancialModellingwithJumpProcesses}
R.~Cont and P.~Tankov.
\newblock \emph{Financial Modelling with Jump Processes}.
\newblock Boca Raton: Chapman \& Hall \/ CRC Press, 2004.

\bibitem[F.~W.~Steutel(2004)]{InfinitelyDivisibleRealLine2004}
K.~v.~Harn F.~W.~Steutel.
\newblock \emph{Infinite Divisibility of Probability Distributions on the Real
  Line}.
\newblock Monogr. Textb. Pure Appl. Math., vol. 259, Marcel Dekker Inc., 2004.

\bibitem[Fukumizu and Leng(2012)]{NIPS2012_1036}
K.~Fukumizu and C.~Leng.
\newblock {Gradient-based kernel method for feature extraction and variable
  selection}.
\newblock In \emph{Annual Conference on Neural Information Processing Systems
  (NIPS)}, pages 2123--2131. 2012.

\bibitem[Fukumizu et~al.(2004)Fukumizu, Bach, and
  Jordan]{DBLP:journals/jmlr/FukumizuBJ03}
K.~Fukumizu, F.~R. Bach, and M.~I. Jordan.
\newblock {Dimensionality Reduction for Supervised Learning with Reproducing
  Kernel Hilbert Spaces}.
\newblock \emph{Journal of Machine Learning Research}, 5:\penalty0 73--99,
  2004.

\bibitem[Fukumizu et~al.(2008)Fukumizu, Gretton, Sun, and
  Sch{\"o}lkopf]{NIPS2007_559}
K.~Fukumizu, A.~Gretton, X.~Sun, and B.~Sch{\"o}lkopf.
\newblock {Kernel Measures of Conditional Dependence}.
\newblock In \emph{Annual Conference on Neural Information Processing Systems
  (NIPS)}, pages 489--496. 2008.

\bibitem[Fukumizu et~al.(2013)Fukumizu, Song, and
  Gretton]{KernelBayes'Rule_BayesianInferencewithPositiveDefiniteKernels}
K.~Fukumizu, L.~Song, and A.~Gretton.
\newblock Kernel bayes' rule: Bayesian inference with positive definite
  kernels.
\newblock \emph{Journal of Machine Learning Research}, pages 3753--3783, 2013.

\bibitem[Gretton et~al.(2008)Gretton, Fukumizu, Teo, Song, Sch{\"o}lkopf, and
  Smola]{Gretton2008independencetest}
A.~Gretton, K.~Fukumizu, C.~H. Teo, L.~Song, B.~Sch{\"o}lkopf, and A.~Smola.
\newblock A kernel statistical test of independence.
\newblock In \emph{Annual Conference on Neural Information Processing Systems
  (NIPS)}. 2008.

\bibitem[Gretton et~al.(2012)Gretton, Borgwardt, Rasch, Sch{\"o}lkopf, and
  Smola]{DBLP:journals/jmlr/GrettonBRSS12}
A.~Gretton, K.~M. Borgwardt, M.~J. Rasch, B.~Sch{\"o}lkopf, and A.~J. Smola.
\newblock {A Kernel Two-Sample Test}.
\newblock \emph{Journal of Machine Learning Research}, 13:\penalty0 723--773,
  2012.

\bibitem[Grosswald(1976)]{SelfDecomposabilityStudentT1976}
E.~Grosswald.
\newblock The student t-distribution of any degree of freedom is infinitely
  divisible.
\newblock \emph{Zeit. Wahrsch. Verw. Gebiete}, 36:\penalty0 103--109, 1976.

\bibitem[Gr{\"u}new{\"a}lder et~al.(2012)Gr{\"u}new{\"a}lder, Lever,
  Baldassarre, Pontil, and Gretton]{DBLP:journals/corr/abs-1206-4655}
S.~Gr{\"u}new{\"a}lder, G.~Lever, L.~Baldassarre, M.~Pontil, and A.~Gretton.
\newblock {Modelling transition dynamics in MDPs with RKHS embeddings}.
\newblock In \emph{International Conference on Machine Learning (ICML)}, pages
  535--542, 2012.

\bibitem[Kanagawa et~al.(2016)Kanagawa, Nishiyama, Gretton, and
  Fukumizu]{KernelMonteCarlo2016}
M.~Kanagawa, Y.~Nishiyama, A.~Gretton, and K.~Fukumizu.
\newblock {Filtering with State-Observation Examples via Kernel Monte Carlo
  Filter}.
\newblock \emph{Neural Computation}, 28:\penalty0 382--444, 2016.

\bibitem[Lee(2010)]{Continuousanddiscretepropertiesofstochasticprocesses}
W.~H. Lee.
\newblock Continuous and discrete properties of stochastic processes.
\newblock \emph{PhD thesis, The University of Nottingham}, 2010.

\bibitem[Lloyd and Ghahramani(2015)]{ModelCriticismNIPS2015}
J.~R. Lloyd and Z.~Ghahramani.
\newblock {Statistical Model Criticism using Kernel Two Sample Test}.
\newblock In \emph{Annual Conference on Neural Information Processing Systems
  (NIPS)}. 2015.

\bibitem[Madan et~al.(1998)Madan, Carr, and
  Chang]{Thevariancegammaprocessandoptionpricing}
B.~D. Madan, P.~Carr, and E.~C. Chang.
\newblock The variance gamma process and option pricing.
\newblock \emph{European Finance Review}, 2:\penalty0 79--105, 1998.

\bibitem[McCalman et~al.(2013)McCalman, O'Callaghan, and Ramos]{McCalman2013}
L.~McCalman, S.~O'Callaghan, and F.~Ramos.
\newblock Multi-modal estimation with kernel embeddings for learning motion
  models.
\newblock In \emph{IEEE International Conference on Robots and Automation
  (ICRA)}, 2013.

\bibitem[Mika et~al.(1999)Mika, Sch{\"o}lkopf, Smola, M{\"u}ller, Scholz, and
  R{\"a}tsch]{Mika99kernelpca}
S.~Mika, B.~Sch{\"o}lkopf, A.~Smola, K.~M{\"u}ller, M.~Scholz, and
  G.~R{\"a}tsch.
\newblock {Kernel PCA and de-noising in feature spaces}.
\newblock In \emph{Annual Conference on Neural Information Processing Systems
  (NIPS)}, pages 536--542, 1999.

\bibitem[Muandet et~al.(2012)Muandet, Fukumizu, Dinuzzo, and
  Sch{\"o}lkopf]{NIPS2012_0015}
K.~Muandet, K.~Fukumizu, F.~Dinuzzo, and B.~Sch{\"o}lkopf.
\newblock {Learning from Distributions via Support Measure Machines}.
\newblock In \emph{Annual Conference on Neural Information Processing Systems
  (NIPS)}, pages 10--18. 2012.

\bibitem[Nishiyama et~al.(2012)Nishiyama, Boularias, Gretton, and
  Fukumizu]{DBLP:conf/uai/NishiyamaBGF12}
Y.~Nishiyama, A.~Boularias, A.~Gretton, and K.~Fukumizu.
\newblock {Hilbert Space Embeddings of POMDPs}.
\newblock In \emph{Uncertainty in Artificial Intelligence (UAI)}, pages
  644--653, 2012.

\bibitem[Nishiyama et~al.(2014)Nishiyama, Kanagawa, Gretton, and
  Fukumizu]{Nishiyama2014}
Y.~Nishiyama, M.~Kanagawa, A.~Gretton, and K.~Fukumizu.
\newblock {Model-based Kernel Sum Rule}.
\newblock In \emph{arXiv: 1409.5178}. 2014.

\bibitem[Nishiyama et~al.(2016)Nishiyama, Afsharinejad, Naruse, Boots, and
  Song]{KernelBayesSmoothingAISTATS2016}
Y.~Nishiyama, A.~H. Afsharinejad, S.~Naruse, B.~Boots, and L.~Song.
\newblock {The Nonparametric Kernel Bayes' Smoother}.
\newblock In \emph{International Conference on Artificial Intelligence and
  Statistics (AISTATS)}. 2016.

\bibitem[Nolan(2013{\natexlab{a}})]{Nolan2013references}
J.~Nolan.
\newblock Bibliography on stable distributions, processes and related topics.
\newblock
  \url{http://citeseerx.ist.psu.edu/viewdoc/download?doi=10.1.1.295.9970&rep=rep1&type=pdf},
  2013{\natexlab{a}}.

\bibitem[Nolan(2013{\natexlab{b}})]{RePEc:spr:compst:v:28:y:2013:i:5:p:2067-2089}
J.~Nolan.
\newblock Multivariate elliptically contoured stable distributions: theory and
  estimation.
\newblock \emph{Computational Statistics}, 28\penalty0 (5):\penalty0
  2067--2089, 2013{\natexlab{b}}.

\bibitem[Prause(1999)]{Prause1999}
K.~Prause.
\newblock \emph{The generalized hyperbolic model: estimation, financial
  derivatives, and risk measures}.
\newblock Ph.D. thesis University of Freiburg, 1999.

\bibitem[Rachev et~al.(2011)Rachev, Kim, Bianchi, and
  Fabozzi]{VolatilityClustering2011}
S.~T. Rachev, Y.~S. Kim, M.~L. Bianchi, and F.~J. Fabozzi.
\newblock \emph{Financial Models with Levy Processes and Volatility
  Clustering}.
\newblock Wiley \& Sons, 2011.

\bibitem[Rahimi and Recht(2007)]{RandomFeaturesNIPS2007}
A.~Rahimi and B.~Recht.
\newblock {Random Features for Large-Scale Kernel Machines}.
\newblock In \emph{Annual Conference on Neural Information Processing Systems
  (NIPS)}. 2007.

\bibitem[Rasmussen and Williams(2006)]{GaussianProcessesforMachineLearning}
C.~E. Rasmussen and C.~K.~I. Williams.
\newblock {Gaussian Processes for Machine Learning}.
\newblock \emph{MIT Press, Cambridge, MA}, 2006.

\bibitem[Rawlik et~al.(2013)Rawlik, Toussaint, and
  Vijayakumar]{PathIntegralControlbyReproducingKernelHilbertSpaceEmbedding2013}
K.~Rawlik, M.~Toussaint, and S.~Vijayakumar.
\newblock {Path Integral Control by Reproducing Kernel Hilbert Space
  Embedding}.
\newblock \emph{International Joint Conference on Artificial Intelligence
  (IJCAI)}, 2013.

\bibitem[Rosi\'nski(2007)]{Rosinski2007}
J.~Rosi\'nski.
\newblock Tempering stable processes.
\newblock \emph{Stochastic Processes and Their Applications}, 117\penalty0
  (6):\penalty0 677--707, 2007.

\bibitem[Samorodnitsky and Taqqu(1994)]{Samorodnitsky1994}
G.~Samorodnitsky and M.~S. Taqqu.
\newblock \emph{Stable non-Gaussian random processes : stochastic models with
  infinite variance}.
\newblock Chapman \& Hall, 1994.

\bibitem[Sato(1999)]{Sato1999}
K.~Sato.
\newblock L\'evy processes and infinitely divisible distributions.
\newblock \emph{Cambridge University Press}, 1999.

\bibitem[Sch\"{o}lkopf and Smola(2002)]{LearningwithKernels2002}
B.~Sch\"{o}lkopf and A.~Smola.
\newblock \emph{Learning with Kernels}.
\newblock MIT Press, Cambridge, 2002.

\bibitem[Schoutens(2003)]{LevyProcessesinFinance_PricingFinancialDerivatives}
W.~Schoutens.
\newblock \emph{L\'evy Processes in Finance: Pricing Financial Derivatives}.
\newblock Chichester: Wiley, 2003.

\bibitem[Smola et~al.(2007)Smola, Gretton, Song, and
  Sch{\"o}lkopf]{Smola_etal_ALT2007}
A.~Smola, A.~Gretton, L.~Song, and B.~Sch{\"o}lkopf.
\newblock {A Hilbert space embedding for distributions}.
\newblock In \emph{International Conference on Algorithmic Learning Theory
  (ALT)}, pages 13--31, 2007.

\bibitem[Song et~al.(2008)Song, Zhang, Smola, Gretton, and
  Sch\"{o}lkopf]{TailoringICML2008}
L.~Song, X.~Zhang, A.~Smola, A.~Gretton, and B.~Sch\"{o}lkopf.
\newblock {Tailoring Density Estimation via Reproducing Kernel Moment
  Matching}.
\newblock \emph{International Conference on Machine Learning (ICML)}, pages
  992--999, 2008.

\bibitem[Song et~al.(2009)Song, Huang, Smola, and
  Fukumizu]{Song+al:icml2010hilbert}
L.~Song, J.~Huang, A.~Smola, and K.~Fukumizu.
\newblock {Hilbert Space Embeddings of Conditional Distributions with
  Applications to Dynamical Systems}.
\newblock In \emph{International Conference on Machine Learning (ICML)}, pages
  961--968, 2009.

\bibitem[Song et~al.(2010)Song, Boots, Siddiqi, Gordon, and Smola]{Song:2010fk}
L.~Song, B.~Boots, S.~M. Siddiqi, G.~J. Gordon, and A.~J. Smola.
\newblock {Hilbert Space Embeddings of Hidden {M}arkov Models}.
\newblock In \emph{International Conference on Machine Learning (ICML)}, pages
  991--998, 2010.

\bibitem[Song et~al.(2011)Song, Gretton, Bickson, Low, and
  Guestrin]{DBLP:journals/jmlr/SongGBLG11}
L.~Song, A.~Gretton, D.~Bickson, Y.~Low, and C.~Guestrin.
\newblock {Kernel Belief Propagation}.
\newblock \emph{Journal of Machine Learning Research - Proceedings Track},
  15:\penalty0 707--715, 2011.

\bibitem[Song et~al.(2013)Song, Fukumizu, and
  Gretton]{KernelEmbeddingofConditionalDistributions}
L.~Song, K.~Fukumizu, and A.~Gretton.
\newblock Kernel embedding of conditional distributions.
\newblock \emph{IEEE Signal Processing Magazine}, 30(4):\penalty0 98--111,
  2013.

\bibitem[Sriperumbudur et~al.(2010)Sriperumbudur, Gretton, Fukumizu, Lanckriet,
  and Sch\"{o}lkopf]{Sriperumbudur_JMLR2010}
B.~Sriperumbudur, A.~Gretton, K.~Fukumizu, G.~Lanckriet, and B.~Sch\"{o}lkopf.
\newblock {Hilbert Space Embeddings and Metrics on Probability Measures}.
\newblock \emph{Journal of Machine Learning Research}, 11:\penalty0 1517--1561,
  2010.

\bibitem[Sriperumbudur et~al.(2011)Sriperumbudur, Fukumizu, and
  Lanckriet]{Sriperumbudur_JMLR2011}
B.~Sriperumbudur, K.~Fukumizu, and G.~Lanckriet.
\newblock {Universality, Characteristic Kernels and RKHS Embedding of
  Measures}.
\newblock \emph{Journal of Machine Learning Research}, 12:\penalty0 2389--2410,
  2011.

\bibitem[Steinwart and Christmann(2008)]{SupportVectorMachines2008}
I.~Steinwart and A.~Christmann.
\newblock \emph{Support Vector Machines}.
\newblock Information Science and Statistics. Springer, 2008.

\bibitem[Thorin(1978)]{AnextensionofthenotionofageneralizedGamma-convolution}
O.~Thorin.
\newblock {An extension of the notion of a generalized $\Gamma$-convolution}.
\newblock \emph{Scandinavian Actuarial Journal}, pages 141--149, 1978.

\bibitem[v.~Hammerstein(2010)]{Hammerstein2010}
E.~A.~F. v.~Hammerstein.
\newblock \emph{Generalized hyperbolic distributions: Theory and applications
  to CDO pricing}.
\newblock Ph.D. thesis University of Freiburg, 2010.

\bibitem[Wendland(2005)]{Wendland2005}
H.~Wendland.
\newblock \emph{Scattered Data Approximation}.
\newblock Cambridge University Press, Cambridge, UK, 2005.

\bibitem[Zolotarev(1986)]{Zolotarev1986}
V.M. Zolotarev.
\newblock \emph{One-dimensional stable distributions}.
\newblock Translations of mathematical monographs, American Mathematical
  Society, 1986.

\end{thebibliography}
\bibliographystyle{unsrt}

\end{document}